\documentclass{vldb}
\usepackage{graphicx}
\usepackage{todonotes,url,subfig,amsmath}
\usepackage[bookmarks=false]{hyperref}
\usepackage{float}
\usepackage{wrapfig}
\usepackage{balance}  % for  \balance command ON LAST PAGE  (only there!)
\usepackage{booktabs}

\restylefloat{figure}

\newcommand{\kgs}{{$k$-NN graphs}}
\newcommand{\kg}{{$k$-NN graph}}
\newcommand{\knn}{{$k$-NN }}
\newcommand{\knnNS}{{$k$-NN}}

\begin{document}

% ****************** TITLE ****************************************

\title{Permutation Search Methods are Efficient, \\ Yet Faster Search is Possible}

\numberofauthors{3} 

\author{
% 1st. author
\alignauthor Bilegsaikhan Naidan\titlenote{Corresponding author.}\\
       \affaddr{Norwegian University of Science and Technology}\\
       \affaddr{Trondheim, Norway}\\
       \email{bileg@idi.ntnu.no}
% 2nd. author
\alignauthor Leonid Boytsov\\
       \affaddr{Carnegie Mellon University}\\
       \affaddr{Pittsburgh, PA, USA}\\
       \email{srchvrs@cs.cmu.edu}\\
% 3rd. author
\alignauthor Eric Nyberg\\
       \affaddr{Carnegie Mellon University}\\
       \affaddr{Pittsburgh, PA, USA}\\
       \email{ehn@cs.cmu.edu}\\
}
\date{9 Jun 2015}

\maketitle

\begin{abstract}
We survey permutation-based methods for approximate $k$-nearest neighbor search.
In these methods, 
every data point is represented by a ranked list of pivots
sorted by the distance to this point.
Such ranked lists are called \emph{permutations}.
The underpinning assumption is that, for both metric and non-metric spaces, the distance 
between permutations is a good proxy for the distance between original points.
Thus, it should be possible to efficiently retrieve most true
nearest neighbors by examining only a tiny subset of data points
whose permutations are similar to the permutation of a query.
We further test this assumption by carrying out an extensive experimental evaluation
where permutation methods are pitted against state-of-the art benchmarks 
(the multi-probe LSH, the VP-tree, and proximity-graph based retrieval)
on a variety of realistically large data set from the image and textual domain.
The focus is on the high-accuracy retrieval methods 
for generic spaces. Additionally, we assume that both data and indices are stored in main memory.
We find permutation methods to be reasonably efficient and
describe a setup where these methods are most useful.
To ease reproducibility, we make our software and data sets publicly available.

\end{abstract}

\section{Introduction}
\emph{Nearest-neighbor searching} is a fundamental operation employed 
in many applied areas including pattern recognition, computer vision,
multimedia retrieval, computational biology, and statistical machine learning.
To automate the search task, real-world objects are represented in a compact
numerical, e.g., vectorial, form and a distance function $d(x,y)$, e.g., 
the Euclidean metric $L_2$, is used to evaluate the similarity of data points $x$ and $y$. 
Traditionally, it assumed that the distance function is 
a non-negative function that is small for similar objects and large for dissimilar one. 
It is equal to zero for identical $x$ and $y$ and is always positive when 
objects are different.

This mathematical formulation allows us to define the nearest-neighbor search
as a \emph{conceptually simple} optimization procedure.
Specifically, given a query data point $q$, the goal is to identify
the nearest (neighbor) data point $x$, i.e.,
the point with the minimum distance value $d(x,q)$ among all data points
(ties can be resolved arbitrarily). 
A natural generalization is a \knn search, where we aim to find $k$ closest
points instead of merely one.
If the distance is not symmetric, two types of queries are considered:
\emph{left} and \emph{right} queries. 
In a \emph{left} query, a data point compared to the query is always
the first (i.e., the left) argument of $d(x,y)$.

Despite being conceptually simple, 
finding nearest neighbors in efficient and effective fashion
is a \emph{notoriously hard} task,
which has been a recurrent topic in the database community (see e.g. \cite{Weber_et_al:1998,Fagin2003,amato2014mi,Liu2014et_al}).
The most studied instance of the problem is an
\emph{exact} nearest-neighbor search in vector spaces, where a distance function is an
actual metric distance (a non-negative, symmetric function satisfying the triangle inequality).
If the search is exact, we must guarantee that 
an algorithm \emph{always} finds a true nearest-neighbor no matter how much computational
resources such a quest may require.
Comprehensive reviews of exact approaches for metric and/or vector spaces can be found
in books by Zezula et al.~\cite{Zezula_et_al:2005} and Samet~\cite{Samet:2005}.

Yet, exact methods work well only in low dimensional metric spaces.\footnote{A dimensionality of a vector space is simply a number of coordinates necessary to represent a vector:
This notion can be generalized to metric spaces without coordinates \cite{chavez2001searching}.}
Experiments showed that exact methods can rarely outperform the sequential scan when dimensionality exceeds ten \cite{Weber_et_al:1998}.
This a well-known phenomenon known as ``the curse of dimensionality''.

Furthermore, a lot of applications are increasingly relying on non-metric spaces 
(for a list of references related to computer vision see, e.g., a work by Jacobs~et~al.~\cite{Jacobs_et_al2000}).
This is primarily because many problems are inherently non-metric \cite{Jacobs_et_al2000}. Thus, using, a non-metric distance permits sometimes a better representation for a domain of interest. 
Unfortunately, exact methods for metric-spaces are not directly applicable to non-metric domains.

Compared to metric spaces, 
it is more difficult to design exact methods for arbitrary non-metric spaces, 
in particular, because they lack sufficiently generic yet simple properties such as the triangle inequality.
When exact search methods for non-metric spaces do exist, they also seem to suffer from the curse of dimensionality \cite{Cayton2008,boytsov2013learning}.

\emph{Approximate} search methods are less affected by the curse of dimensionality \cite{Pestov:2012}
and can be used in various non-metric spaces when exact retrieval is not necessary \cite{skopal2007,goh2002dyndex,chen2008efficient,Cayton2008,boytsov2013learning}.
Approximate search methods can be much more efficient than exact ones, 
but this comes at the expense of a reduced search accuracy.
The quality of approximate searching is often measured using \emph{recall},
which is equal to the average fraction of true neighbors returned by a search method.
For example, if the method routinely misses every other true neighbor, the respective recall
value is 50\%. 

Permutation-based algorithms is 
an important class of approximate retrieval methods
that was independently introduced by Amato~\cite{Amato:2008} 
and Ch\'{a}vez et al.~\cite{Chavez_et_al:2008}.
It is based on the idea that if we rank a set of reference points--called
\emph{pivots}--with respect to distances from a given point,
the pivot rankings produced by two near points should be similar.
A number of methods based on this idea 
were recently proposed and evaluated \cite{Amato:2008,Chavez_et_al:2008,esuli2009pp,chavez2015near,amato2014mi}
(these methods are briefly surveyed in \S~\ref{SectionLitSurvey}).
However, a comprehensive evaluation that involves a
diverse set of large metric and non-metric data sets (i.e.,
asymmetric and/or hard-to-compute distances) is lacking.
In \S~\ref{SectionExperiments},
we fill this gap by carrying out an extensive experimental evaluation 
where these methods (implemented by us) are compared against some 
of the most efficient state-of-the art benchmarks.
The focus is on the high-accuracy retrieval methods  (recall close to 0.9) for generic spaces.
Because distributed high-throughput main memory databases are gaining popularity (see., e.g. \cite{Kallman2008}),
we focus on the case where data and indices are stored in main memory.
Potentially, the data set can be huge, yet, we run experiments
only with a smaller subset that fits into a memory of one server. 

%The acknowledgments are published separately.\footnote{\url{http://arxiv.org/abs/1506.03163}}

\section{Permutation Methods}\label{SectionLitSurvey}
\subsection{Core Principles}
Permutation methods are \emph{filter-and-refine} methods belonging to the class of pivoting searching techniques.
\emph{Pivots} (henceforth denoted as $\pi_i$) are reference points randomly selected during indexing. 
To create an index, we compute the distance from every data point $x$ to every pivot $\pi_i$. 
We then memorize either the original distances or some distance statistics
in the hope that these statistics can be useful during searching.
At search time, we compute distances from the query to pivots and 
prune data points using, e.g., the triangle inequality \cite{Samet:2005,Zezula_et_al:2005}
or its generalization for non-metric spaces \cite{farago1993fast}.

\begin{figure}[t]
\centering\includegraphics[width=0.20\textwidth]{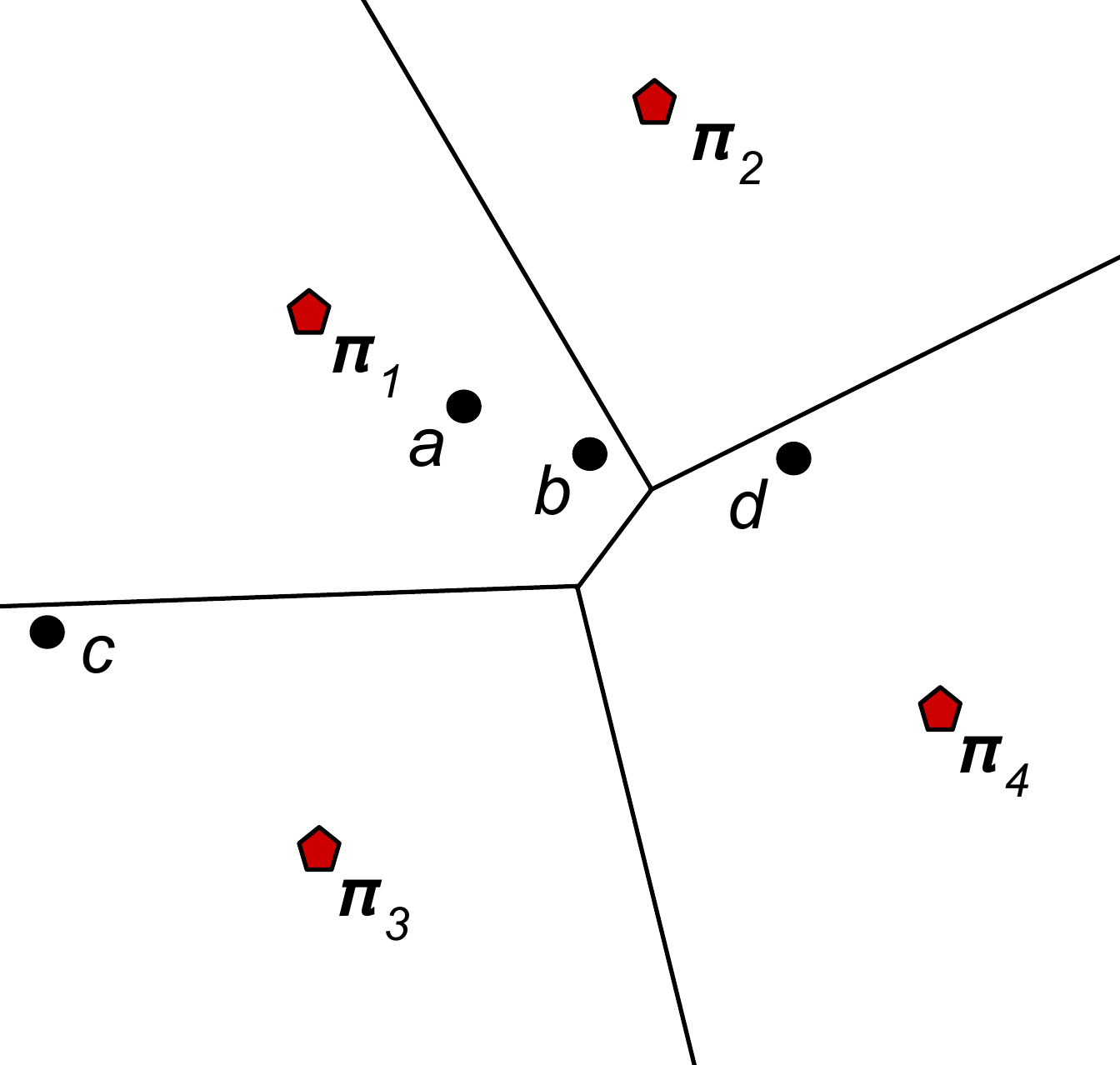}
\caption{\label{FigExample}Voronoi diagram produced by four pivots $\pi_i$.
The data points are $a$, $b$, $c$, and $d$. The distance is $L_2$.}
\end{figure}

Alternatively, 
rather than relying on distance values directly,
we can use precomputed statistics to produce estimates for distances between the query and data points. 
In particular, in the case of permutation methods, 
we assess similarity of objects based on their relative distances to pivots.
To this end, for each data point $x$,
we arrange pivots $\pi_i$ in the order of increasing distances from $x$.
The ties can be resolved, e.g., by selecting a pivot with the smallest index.
Such a \emph{permutation} (i.e., ranking) of pivots is essentially a vector whose \mbox{$i$-th}
element keeps an ordinal position of the \mbox{$i$-th} pivot in the set of pivots
sorted by their distances from $x$.
We say that point $x$ \emph{induces} the permutation.

Consider the Voronoi diagram in Figure~\ref{FigExample} produced by pivots $\pi_1$, $\pi_2$, $\pi_3$, and $\pi_4$.
Each pivot $\pi_i$ is associated with its own cell containing points  that are \emph{closer} to $\pi_i$ than
to any other pivot $\pi_j, i\ne j$. 
The neighboring cells of two pivots are separated by a segment of the line \emph{equidistant} to these pivots.
Each of the data points $a$, $b$, $c$, and $d$ ``sits'' in the cell of its
closest pivot. 

For the data point $a$, points $\pi_1$, $\pi_2$, $\pi_3$, and $\pi_4$ are respectively the first, the second,
the third, and the forth closest pivots. Therefore, the point $a$ induces the permutation $(1,2,3,4)$.
For the data point $b$, which is the nearest neighbor of $a$, two closest pivots are also $\pi_1$ and $\pi_2$. 
However, $\pi_4$ is closer than $\pi_3$. Therefore, the permutation induced by $b$ is $(1,2,4,3)$.
Likewise, the permutations induced by $c$ and $d$ are 
$(2, 3, 1, 4)$ and $(3, 2, 4, 1)$, respectively.

The \emph{underpinning assumption} of permutation methods is that most nearest neighbors
can be found by retrieving a small fraction of data points whose pivot rankings, 
i.e., the induced permutations, are similar to the pivot ranking of the query.
Two most popular choices to compare the rankings $x$ and $y$ are:
Spearman's rho distance (equal to the squared $L_2$) 
and the Footrule distance (equal to $L_1$) \cite{diaconis1988group,Chavez_et_al:2008}. More formally,
$\mbox{SpearmanRho}(x,y)=\sum_i (x_i-y_i)^2$ and $\mbox{Footrule}(x,y) =\sum_i |x_i-y_i|$.
According to Ch\'{a}vez et al.~\cite{Chavez_et_al:2008}  Spearman's rho 
is more effective than the Footrule distance. This was also confirmed by our own experiments.

Converting the vector of distances to pivots into a permutation entails information loss,
but this loss is not necessarily detrimental.
In particular, our preliminary experiments showed 
that using permutations instead of vectors of original distances results in slightly better retrieval performance.
The information about relative positions of the pivots can be further coarsened by binarization:
All elements smaller than a threshold $b$ become zeros and elements at least as large as $b$ become ones \cite{tellez2009brief}.
The similarity of binarized permutations is computed via the Hamming distance.

In the example of Figure~\ref{FigExample}, the values of the Footrule distance
between the permutation of $a$ and permutations of $b$, $c$, and $d$ are
equal to 
%(1,2,3,4)-(1,2,3,4) =0+0+1+1=2
%(1,2,3,4)-(2,3,1,4) =1+1+2+0=4
%(1,2,3,4)-(3,2,4,1) =2+0+1+3=6
2, 4, and 6, respectively.
Note that the Footrule distance on permutations correctly ``predicts'' the closest neighbor of $a$.
Yet, the ordering of points based on the Footrule distance is not perfect: the Footrule distance from the permutation of $a$ to 
the permutation of its second nearest neighbor $d$ is larger than the Footrule distance to the permutation of the third nearest neighbor $c$.

Given the threshold $b=3$, the binarized permutations induced by $a$, $b$, $c$, and $d$ are equal to
$(0,0,1,1)$, $(0,0,1,1)$, $(0,1,0,1)$, and $(1,0,1,0)$, respectively.
In this example, the binarized permutation of $a$ and its nearest neighbor $b$ are equal, i.e.,
the distance between respective permutations is zero. 
When we compare $a$ to $c$ and $d$, the Hamming distance does not discriminate between $c$ and $d$ as
their binary permutations are both at distance two from the binary permutation of $a$.

Permutation-based searching belongs to a class of \emph{filter-and-refine} methods, 
where objects are mapped to data points in a low-dimensional space (usually $L_1$ or $L_2$).
%Usually this low dimensional space is $L_1$ or $L_2$.
Given a permutation of a query, we carry out a nearest neighbor search in
the space of permutations. 
Retrieved entries represent a (hopefully) small list of candidate data
points that are compared directly to the query using the distance 
in the \emph{original} space.
The permutation methods differ in ways of producing candidate records, i.e., in the way of carrying out
the filtering step.
%Major approaches include brute-force searching in the space of permutations
%as well as indexing methods that rely on either prefix-indexing, inverted files, or classic
%data structures for metric spaces.
In the next sections we describe these methods in detail.

Permutation methods are similar to the rank-aggregation method OMEDRANK due to Fagin et~al.~\cite{Fagin2003}.
In OMEDRANK there is a small set of voting pivots,
each of which ranks data points based on a somewhat imperfect notion of the distance from points to the query (e.g., computed
via a random projection).
While each individual ranking is imperfect, 
a more accurate ranking can be achieved by rank aggregation.
Thus, unlike permutation methods, OMEDRANK uses pivots to rank data points 
and aims to find an \emph{unknown} permutation of \emph{data points} that 
reconciles differences in data point rankings in the best possible way.
When such a consolidating ranking is found, the most highly ranked objects from this
\emph{aggregate} ranking can be used as answers to a nearest-neighbor query.
Finding the aggregate ranking is an NP-complete problem that Fagin~et~al.~\cite{Fagin2003} solve only heuristically.
In contrast, permutation methods use data points to rank pivots and solve a much simpler problem of finding \emph{already known and computed} permutations of \emph{pivots} that are the best matches for the query permutation. 

\begin{table*}
\centering
\caption{\label{TableDataSet}Summary of Data Sets}
{
\begin{tabular}{l c c c  c  c  l }
\toprule
{Name}       & {Distance} & {\# of rec.} & {Brute-force} & {In-memory}  & {Dimens.}     &  {Source} \\
                    & {function} & &{search (sec)} & {size}  &                          &  \\
\midrule
\multicolumn{7}{c}{\textbf{Metric Data}}\\
\midrule
CoPhIR              & $L_2$     & $5\cdot10^6$ &  0.6 & 5.4GB  & 282 & MPEG7 descriptors \cite{CoPhIR} \\ 
SIFT                & $L_2$     & $5\cdot10^6$ &  0.3 & 2.4GB  & 128 & SIFT descriptors \cite{jegou2011searching} \\ 
ImageNet            & SQFD\cite{Beecks:2013} 
                                & $1\cdot10^6$ &  4.1 & 0.6 GB & N/A  & Signatures generated from \\
                    &           &             &  &         &       &    ImageNet LSVRC-2014~\cite{ILSVRCarxiv14} \\
\midrule
\multicolumn{7}{c}{\textbf{Non-Metric Data}}\\
\midrule
Wiki-sparse & Cosine sim. & $4\cdot 10^6$ & 1.9  & 3.8GB & $10^5$ &                    Wikipedia TF-IDF vectors \\
                 &  &                &       &        &    &  generated via Gensim \cite{GENSIM}        \\
Wiki-8  & KL-div/JS-div  & $2\cdot 10^6$ & 0.045/0.28 & 0.13GB & 8 &  LDA (8 topics) generated \\
                &            & & & & &  from Wikipedia via Gensim \cite{GENSIM} \\
Wiki-128   & KL-div/JS-div  & $2\cdot 10^6$ & 0.22/4 & 2.1GB & 128 &  LDA (128 topics) generated \\
                &            & & & & &  from Wikipedia via Gensim \cite{GENSIM} \\
DNA           & Normalized & $1\cdot10^6$ & 3.5  & 0.03GB & N/A & Sampled from the Human Genome\footnote{\url{http://hgdownload.cse.ucsc.edu/goldenPath/hg38/bigZips/}} \\   
              & Levenshtein &              &       &    &  & with sequence length $ \mathcal{N}(32,4)$     \\ 
\bottomrule
\end{tabular}
}
\end{table*}

\subsection{Brute-force Searching of Permutations}
In this approach, the filtering stage is implemented as a brute-force comparison of the query permutation
against the permutations of the data with subsequent selection of the $\gamma$ entries
that are $\gamma$-nearest objects in the space of permutations.
A number of candidate entries $\gamma$ is a parameter of the search algorithm that is
often understood as a fraction (or percentage) of the total number of points.
Because the distance in the space of permutations
is not a perfect proxy for the original distance,
to answer a \knnNS-query with high accuracy,
the number of candidate records has to be much larger than $k$ (see \S~\ref{SectionIndexability}).

A straightforward implementation of brute-force searching relies on a priority queue. 
Ch\'{a}vez et al.~\cite{Chavez_et_al:2008} proposed to use incremental sorting as a more efficient alternative.
In our experiments with the $L_2$ distance, the latter approach is twice as fast as the approach relying on a standard C++ implementation of a priority queue.

The cost of the filtering stage can be reduced by using binarized permutations \cite{tellez2009brief}.
Binarized permutations can be stored compactly as bit arrays.
Computing the Hamming distance between bit arrays
can be done efficiently by
XOR-ing corresponding computer words and counting the number of non-zero bits of the result.
For bit-counting, one can use a special instruction available on many modern CPUs.
\footnote{In C++, this instruction is provided via the intrinsic function \texttt{\_\_builtin\_popcount}.}

The brute-force searching in the permutation space, unfortunately, is not very efficient,
especially if the distance can be easily computed:
If the distance is ``cheap'' (e.g., $L_2$) and the index is stored in main memory,
the brute-force search in the space of permutations is not much faster than the brute-force search in the original space.

\subsection{Indexing of Permutations}
To reduce the cost of the filtering stage of permutation-based searching, 
three types of indices were proposed: 
the Permutation Prefix Index (PP-Index) \cite{esuli2009pp}, 
existing methods for metric spaces \cite{figueroa2009speeding},
and the Metric Inverted File (MI-file) \cite{Amato:2008}. 

Permutations are integer vectors whose values are between one and the total number of pivots $m$.
We can view these vectors as sequences of symbols over a finite alphabet and index these sequences using a prefix tree. 
This approach is implemented in the PP-index. 
At query time, the method aims to retrieve $\gamma$ candidates
by finding permutations that share a prefix of a given length with the permutation
of the query object.
This operation can be carried out efficiently via the prefix tree constructed
at index time.
If the search generates fewer candidates than a specified threshold $\gamma$, 
the procedure is recursively repeated using a shorter prefix.
For example, the permutations of points $a$, $b$, $c$, and $d$ in Figure~\ref{FigExample} can be seen
as strings \texttt{1234}, \texttt{1243}, \texttt{2314}, and \texttt{3241}.
The permutation of points $a$ and $b$, which are nearest neighbors, share a two-character prefix with $a$.
In contrast, permutations of points $c$ and $d$, which are more distant from $a$ than $b$, have no common prefix with $a$.

To achieve good recall, it may be necessary to use short prefixes.
However, longer prefixes are more selective 
than shorter ones (i.e., they generate fewer candidate records)
and are, therefore, preferred for efficiency reasons.
In practice, a good trade-off between recall and efficiency
is typically achieved only by building several copies of the PP-index (using
different subsets of pivots) \cite{amato2014mi}. 

Figueroa and Fredriksson experimented with indexing permutations using
well-known data structures for metric spaces \cite{figueroa2009speeding}.
Indeed, the most commonly used permutation distance: Spearman's rho,
is a monotonic transformation (squaring) of the Euclidean distance.
Thus, it should be possible to find $\gamma$ nearest neighbors by indexing
permutations, e.g., in a VP-tree \cite{yianilos1993data,uhlmann1991satisfying}.

Amato and Savino proposed to index permutation using an inverted file \cite{Amato:2008}.
They called their method the MI-file.
To build the MI-file, they first select $m$ pivots
and compute their permutations/rankings induced by data points.
For each data point, $m_i \le m$ most closest pivots are indexed in the inverted file.
Each posting is a pair $(pos(\pi_i, x), x)$, where
$x$ is the identifier of the data point and $pos(\pi_i,x)$ is a position
of the pivot in the permutation induced by $x$.
Postings of the same pivot are sorted by pivot's positions.

Consider the example of Figure~\ref{FigExample} and imagine that we index two most closest pivots (i.e., $m_i=2$).
The point $a$ induces the permutation $(1,2,3,4)$. Two closest pivots $\pi_1$ and $\pi_2$ generate
postings $(1,a)$ and $(2,a)$. The point $b$ induces the permutation $(1,2,4,3)$. Again, $\pi_1$ and $\pi_2$
are two pivots closest to $b$. The respective postings are $(1,b)$ and $(2,b)$.
The permutation of $c$ is $(2,3,1,4)$. Two closest pivots are $\pi_1$ and $\pi_3$. The respective postings are
$(2,c)$ and $(1,c)$. The permutation of $d$ is $(3,2,4,1)$. Two closest pivots are $\pi_2$ and $\pi_4$ with
corresponding postings $(2,d)$ and $(1,d)$.

At query time, we select $m_s \le m_i$ pivots closest to the query $q$ and retrieve respective posting lists.
If $m_s = m_i = m$, it is possible to compute the exact Footrule distance
(or Spearman's rho) between the query permutation and the permutation 
induced by data points. 
One possible search algorithm keeps an accumulator (initially set to zero) for every data point.
Posting lists are read one by one:
For every encountered posting $(pos(\pi_i,x), x)$
 we increase the accumulator of $x$ by the value $|pos(\pi_i,x)-pos(\pi_i,q)|$.
If the goal is to compute Spearman's rho, the accumulator is increased by
$|pos(\pi_i,x)-pos(\pi_i,q)|^2$.

If $m_s < m$, by construction of the posting lists, using the inverted index,
it is possible to obtain rankings of only $m_s < m$ pivots.
For the remaining, $m-m_s$ pivots we pessimistically assume that their rankings
are all equal to $m$ (the maximum possible value).
Unlike the case $m_i=m_s=m$, all accumulators are initially set to $m_s \cdot m$.
Whenever we encounter a posting posting $(pos(\pi_i,x), x)$
 we subtract  $m - |pos(\pi_i,x)-pos(\pi_i,q)|$ from the accumulator of $x$.

Consider again the example of Figure~\ref{FigExample}. Let  $m_i=m_s=2$ and
$a$ be the query point. Initially, the accumulators of $b$, $c$, and $d$ contain values $4 \cdot 2=8$.
Because $m_s=2$, we read posting lists only of the two closest pivots for the query point $a$, i.e., $\pi_1$
and $\pi_2$. The posting lists of $\pi_1$ is comprised of $(1,a)$, $(1,b)$, and $(2,c)$.
On reading them (and ignoring postings related to the query $a$),
 accumulators $b$ and $c$ are decreased by $4-|1-1|=4$ and $4-|1-2|=3$, respectively.
The posting lists of $\pi_2$ are $(2,a)$, $(2,b)$, and $(2,d)$.
On reading them, we subtract $4-|2-2|=4$ from each of the accumulators $b$ and $d$.
In the end, the accumulators $b$, $c$, $d$ are equal to 0, 5, and 4.
Unlike the case when we compute the Footrule distance between complete permutation, 
the Footrule distance on truncated permutations correctly predicts the order of three nearest neighbors of $a$.

Using fewer pivots at retrieval time allows us to reduce the number of processed 
posting lists. Another optimization consists in keeping posting lists sorted
by pivots position $pos(\pi_i, x)$ and retrieving only the entries
satisfying the following restriction on the maximum position difference: $|pos(\pi_i, x) - pos(\pi_i, q)| \le D$, where $D$ is a method parameter.
Because posting list entries are sorted
by pivot positions, the first and the last entry satisfying the maximum
position difference requirement can be efficiently found via the binary search.

Tellez~et~al.~\cite{tellez2013succinct} proposed a modification of the
MI-file which they called a Neighborhood APProximation index (NAPP).
In the case of NAPP, there also exist a large set of $m$ pivots
of which only $m_i < m$ pivots (most closest to inducing data points)
are indexed.
Unlike the MI-file, however, posting lists contain only object identifiers,
but no positions of pivots in permutations.
Thus, it is not possible to compute an estimate for the Footrule distance
by reading only posting lists.
Therefore, instead of an estimate for the Footrule distance,
the number of most closest \emph{common} pivots is used to sort candidate objects.
In addition, the candidate objects sharing with the query fewer than $t$ 
closest pivots are discarded ($t$ is a parameter).
For example, points $a$ and $b$ in Figure~\ref{FigExample} share the same common pivot $\pi_1$.
At the same time $a$ does not share any closest pivot with points $d$ and $c$. 
Therefore,
if we use $a$ as a query, the point $b$ will be considered to be the best candidate point.

Ch\'{a}vez et al.~\cite{Chavez_et_al:2015} proposed a single framework that unifies several approaches
including PP-index, MI-file, and NAPP. Similar to the PP-index,
permutations are viewed as strings over a finite alphabet.
However, these strings are indexed using a special sequence index (rather than a prefix tree) that efficiently
supports rank and select operations. These operations can be used to simulate
various index traversal modes, including, e.g., retrieval of all strings whose
\mbox{$i$-th} symbol is equal to a given one.

\section{Experiments}\label{SectionExperiments}
\subsection{Data Sets and Distance Functions}
We employ three image data sets: CoPhIR, SIFT, ImageNet, 
and several data sets created from textual data.
The smallest data set (DNA) has one million entries, 
while the largest one (CoPhIR)
contains five million high-dimensional vectors. 
%When CoPhIR is indexed using a multi-probe LSH (\S~\ref{SectionMethods}), we
%consume 19GB out of 32GB available on our server.
All data sets derived from Wikipedia were generated using the topic modelling library GENSIM \cite{GENSIM}. 
The data set meta data is summarized 
in Table~\ref{TableDataSet}. Below, we describe our data sets in detail.

\textbf{CoPhIR} is a five million subset of MPEG7 descriptors downloaded
from the website of the Institute of the National Research Council of Italy\cite{CoPhIR}.

\textbf{SIFT} is a five million subset of SIFT descriptors (from the learning subset) downloaded from a TEXMEX collection website\cite{jegou2011searching}.\footnote{\url{http://corpus-texmex.irisa.fr/}} 

In experiments involving CoPhIR and SIFT, we employed $L_2$
to compare unmodified, i.e., raw visual descriptors. We implemented 
an optimized procedure to compute $L_2$
that relies on Single Instruction Multiple Data (SIMD) operations available on Intel-compatible CPUs.  Using this implementation, it is possible to carry out about 30 million 
$L_2$ computations per second using SIFT vectors or 10 million $L_2$
computations using CoPhIR vectors.

\textbf{ImageNet} collection comprises one million signatures extracted from 
LSVRC-2014 data set~\cite{ILSVRCarxiv14}, which contains 1.2 million high resolution 
images.
We implemented our own code to extract signatures following the method of Beecks~\cite{Beecks:2013}.
For each image, we selected $10^4$ pixels randomly and 
mapped them into 7-dimensional feature space: 
three color, two position, and two texture dimensions.

The features were clustered by the standard $k$-means algorithm with 20 clusters.
Then, each cluster was represented by an 8-dimensional vector, which included
a 7-dimensional centroid and a cluster weight (the number of cluster points divided
by $10^4$).

\begin{table*}
\footnotesize
\centering
\caption{\label{TableIndexSizeTime}Index Size and Creation Time for Various Data Sets}
\begin{tabular}{l r@{\hspace{1mm}}r r@{\hspace{1mm}}r r@{\hspace{1mm}}r r@{\hspace{1mm}}r r@{\hspace{1mm}}r r@{\hspace{1mm}}r }
\toprule
               & \multicolumn{2}{c}{VP-tree}
               & \multicolumn{2}{c}{NAPP}
               & \multicolumn{2}{c}{LSH}
               & \multicolumn{2}{c}{Brute-force filt.}
               & \multicolumn{2}{c}{\knn graph}\\
\midrule
\multicolumn{11}{c}{\textbf{Metric Data}}\\
\midrule        %    vp-tree         napp               lsh                  perm.f         knngraph
CoPhIR             & 5.4 GB & (0.5min)  & 6 GB  & (6.8min)      & 13.5 GB &(23.4min)   &         &          & 7 GB    & (52.1min)  
\\
SIFT               & 2.4 GB    & (0.4min)  & 3.1 GB & (5min)      & 10.6 GB & (18.4min)     &         &          & 4.4 GB  &(52.2min) 
\\
ImageNet           & 1.2 GB  & (4.4min)  & 0.91 GB  & (33min)     &      &             & 12.2 GB & (32.3min)  & 1.1 GB  &(127.6min) 
\\
\midrule
\multicolumn{11}{c}{\textbf{Non-Metric Data}}\\
\midrule
Wiki-sparse        &         &           & 4.4 GB &   (7.9min)    &   &     &      &      & 5.9 GB & (231.2min)
\\
Wiki-8 (KL-div)    & 0.35 GB  & (0.1min)    & 0.67 GB & (1.7min)    &    &     &       &      & 962 MB & (11.3min)  
\\
Wiki-128 (KL-div)  & 2.1 GB  & (0.2min)    & 2.5 GB  & (3.1min)    &    &     &      &      & 2.9 GB & (14.3min)   
\\
Wiki-8 (JS-div)    & 0.35 GB  & (0.1min)    & 0.67 GB & (3.6min)    &    &      &       &     &  2.4 GB & (89min)
\\
Wiki-128 (JS-div)  & 2.1 GB & (1.2min)    & 2.5 GB & (36.6min)   &    &       &       &     & 2.8 GB & (36.1min)    
\\
DNA                & 0.13 GB & (0.9min)    & 0.32 GB & (15.9min)    &    &      & 61 MB & (15.6min) &  1.1 GB & (88min)
\\
\midrule
\multicolumn{11}{l}{\textbf{Note:} The indexing algorithms of NAPP and \knn graphs used four threads.}\\
\multicolumn{11}{l}{In all but two cases (DNA and Wiki-8 with JS-divergence), we build the \knn graph using the Small World algorithm \cite{malkov2014approximate}.}\\
\multicolumn{11}{l}{In the case of  DNA or Wiki-8 with JS-divergence, we build the \knn graph using the NN-descent algorithm \cite{dong2011efficient}.}\\
\bottomrule
\end{tabular}
\end{table*}

\begin{figure*}
\centering
\subfloat[\small SIFT ($L_2$) \textbf{rnd-proj}\label{CorrRandL2}]{\hspace{-1em}\includegraphics[width=0.27\textwidth]{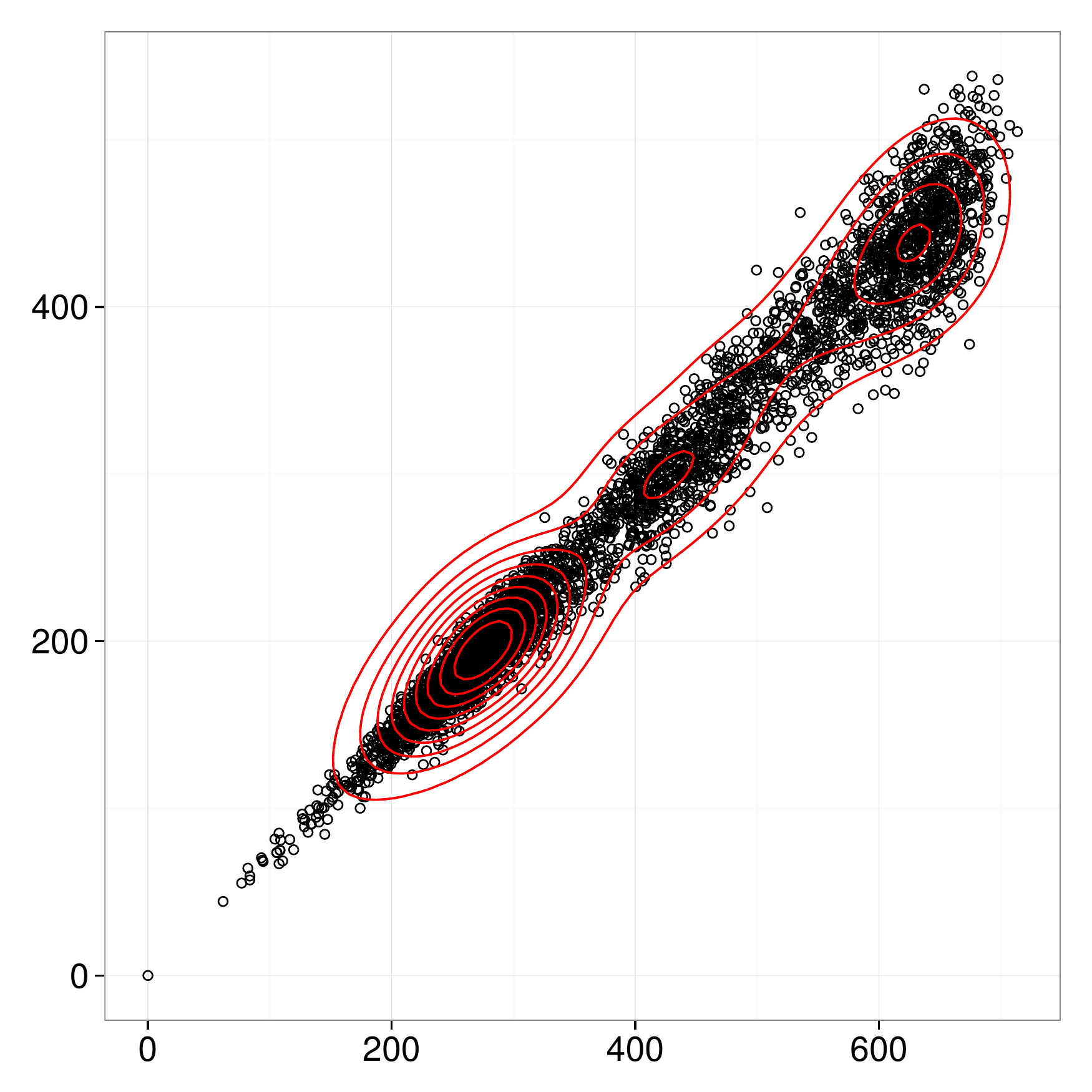}\hspace{-1em}}
\subfloat[\small Wiki-sparse (cos.) \mbox{\textbf{rnd-proj}}\label{CorrRandCos}]{\includegraphics[width=0.27\textwidth]{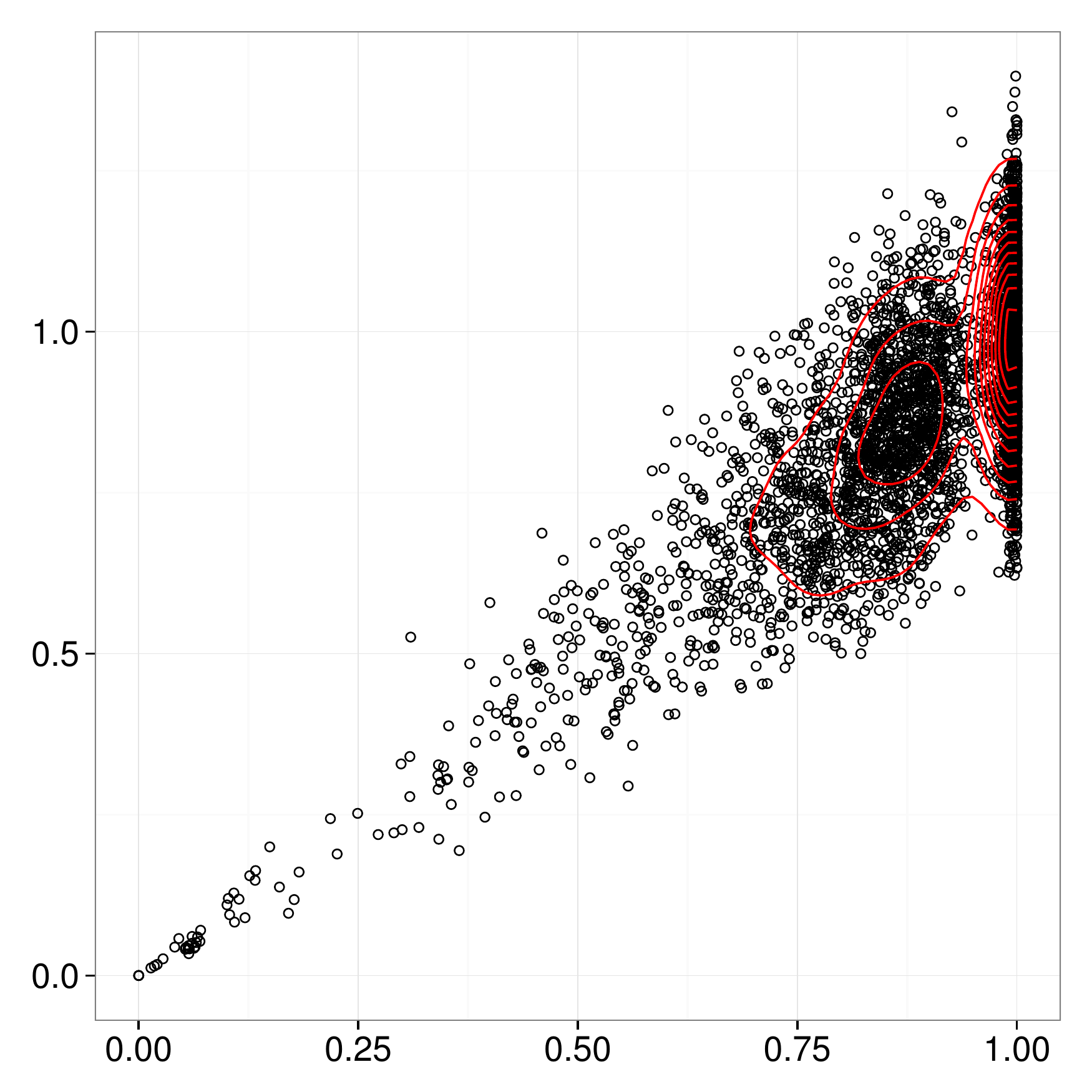}\hspace{-1em}}
\subfloat[\label{UnkProj2}\small Wiki-8 \mbox{(KL-div)} \textbf{perm}]{\includegraphics[width=0.27\textwidth]{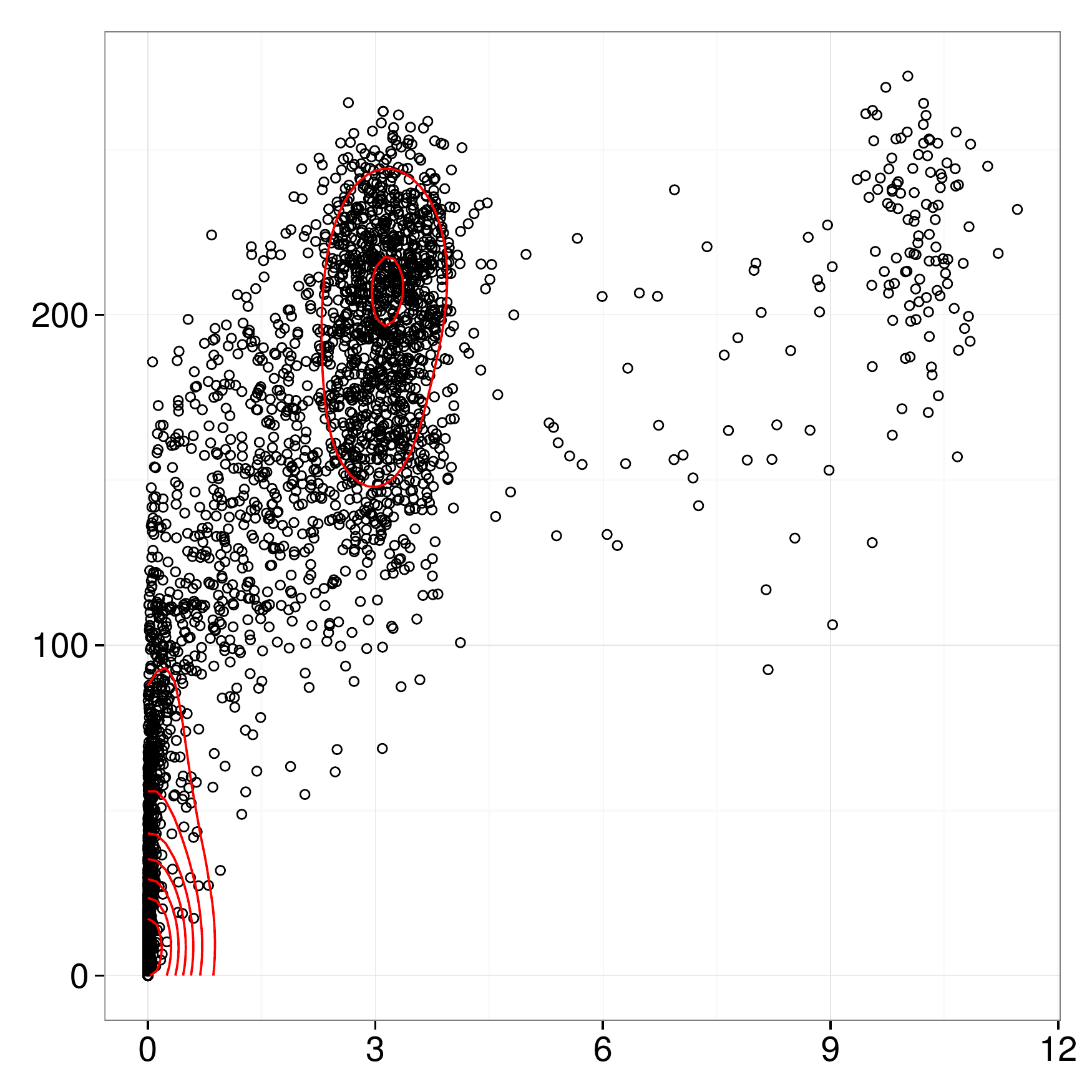}\hspace{-1em}}
\subfloat[\label{GoodProj3}\small DNA (norm. Leven.) \textbf{perm}]{\includegraphics[width=0.27\textwidth]{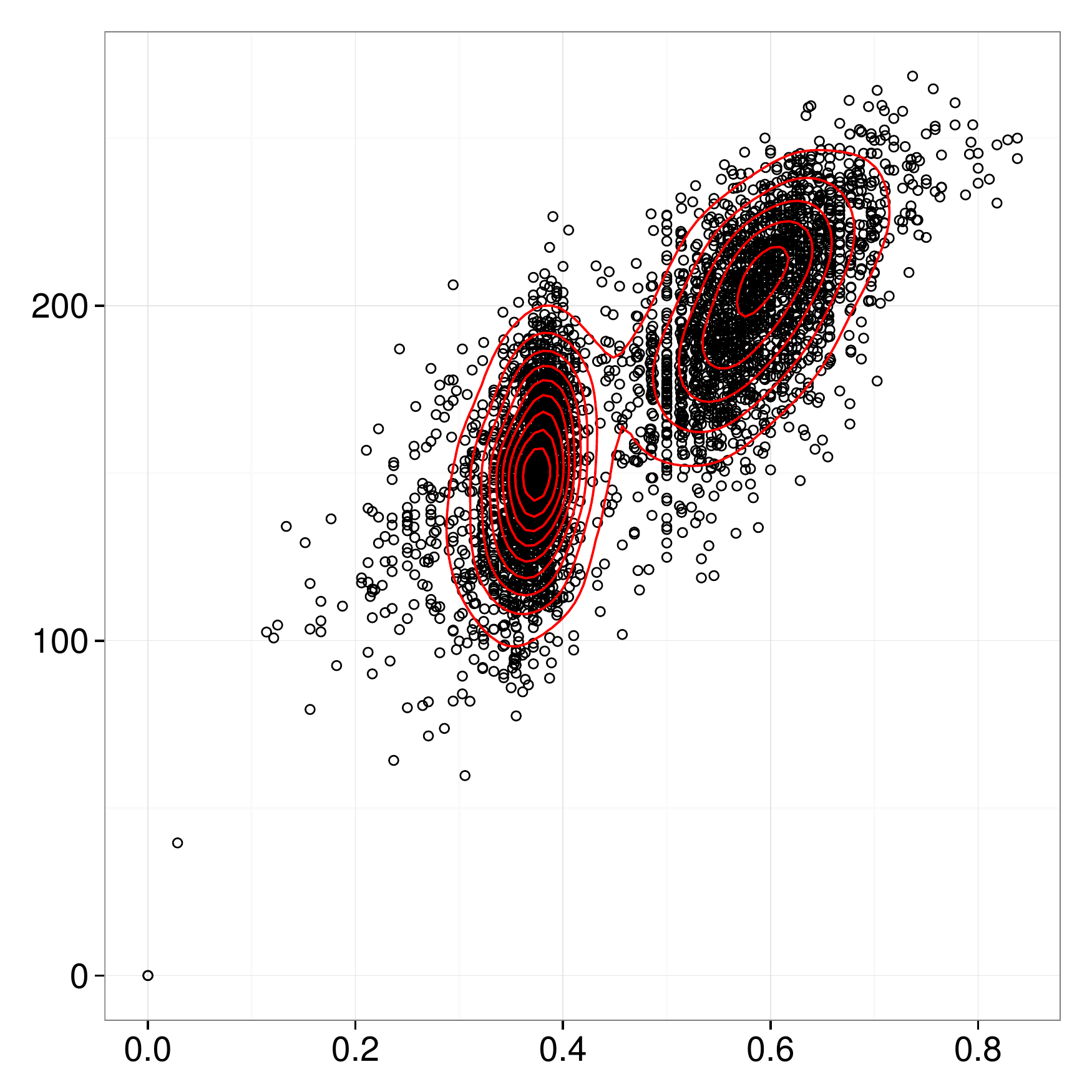}\hspace{-1em}}
\\
\subfloat[\label{GoodProj1}\small SIFT ($L_2$) \textbf{perm}]{\hspace{-1em}\includegraphics[width=0.27\textwidth]{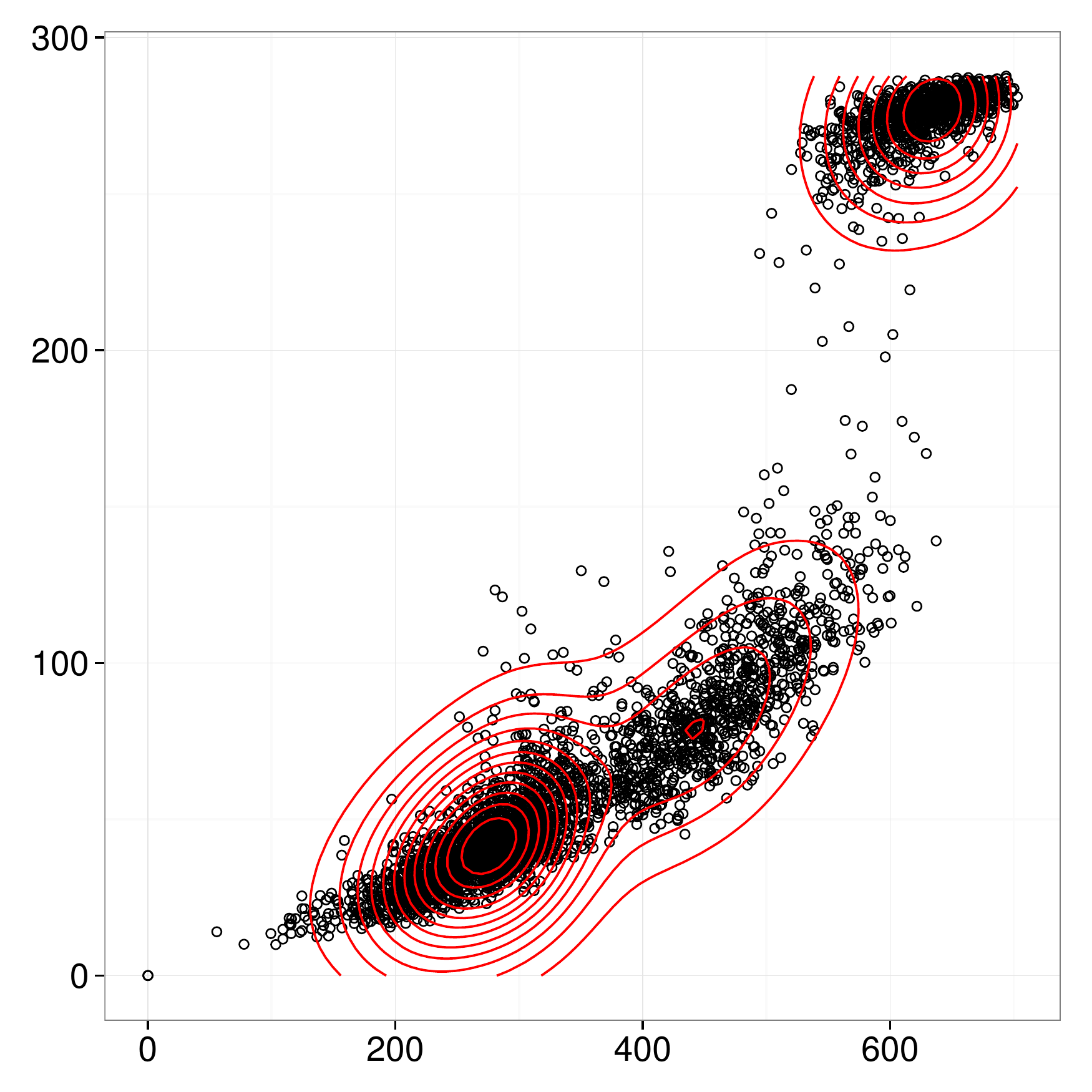}\hspace{-1em}}
\subfloat[\label{UnkProj1}\small Wiki-sparse  \textbf{perm}]{\includegraphics[width=0.27\textwidth]{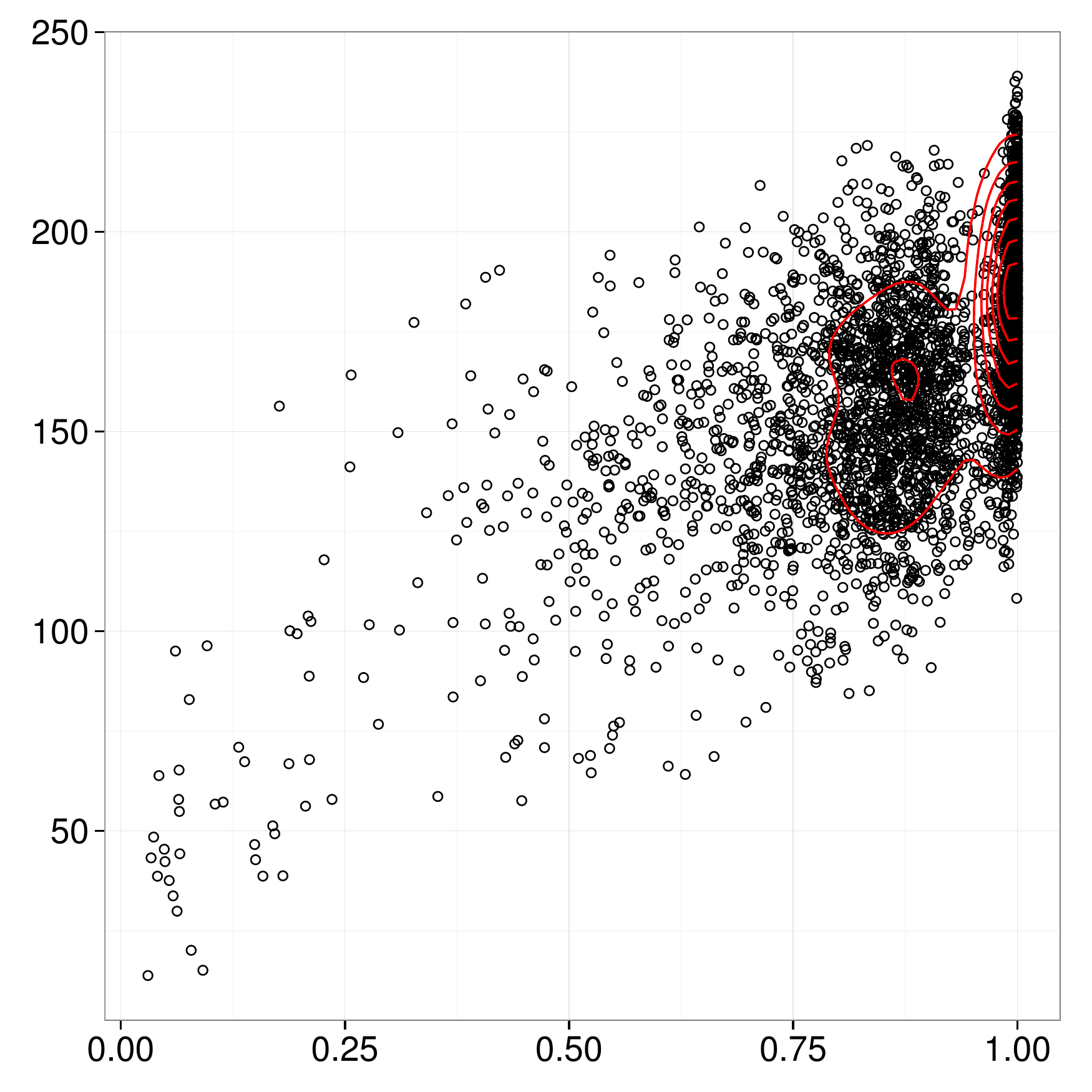}\hspace{-1em}}
\subfloat[\label{BadProj1}\small Wiki-128 \mbox{(KL-div)} \textbf{perm}]{\includegraphics[width=0.27\textwidth]{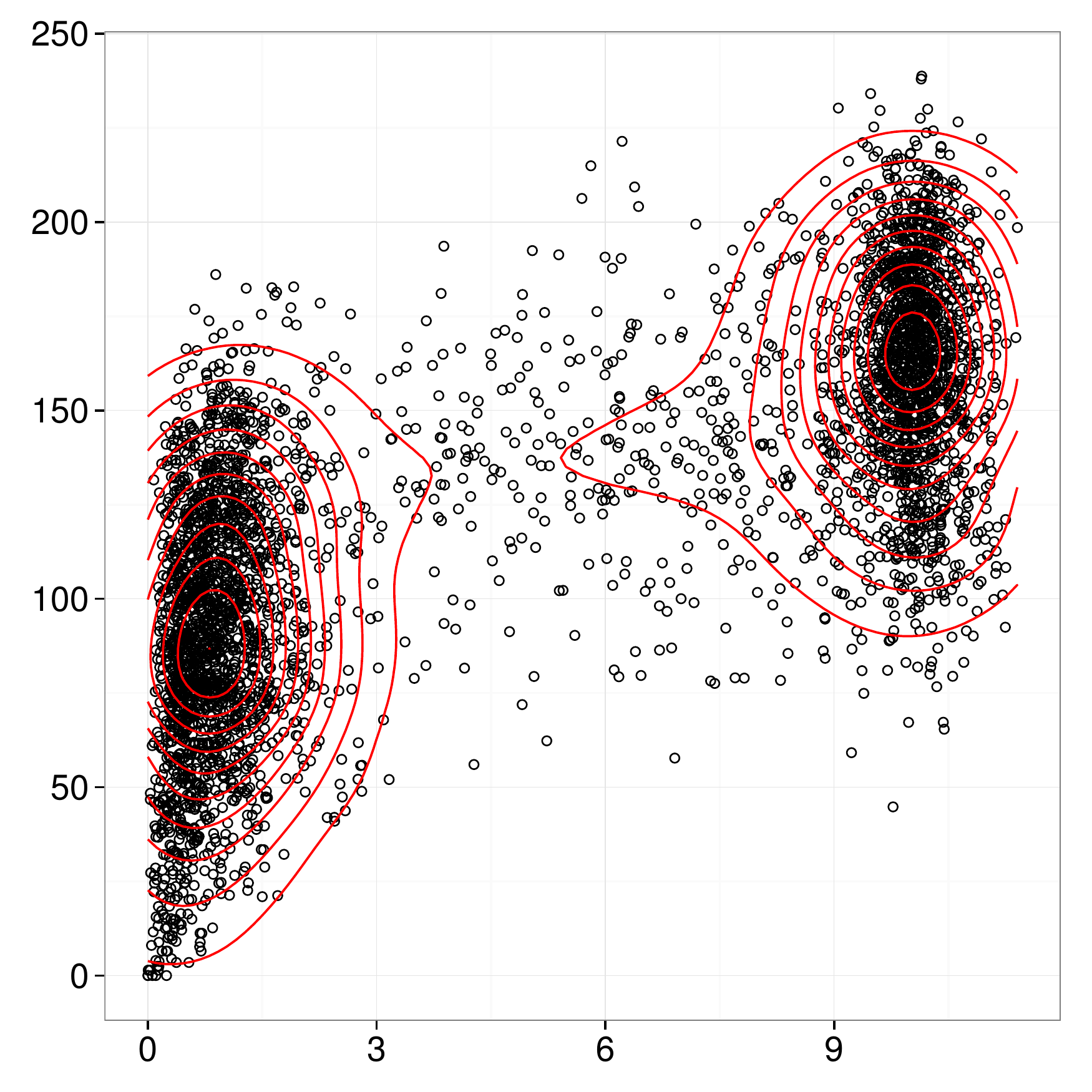}}
\subfloat[\label{GoodProj2}\small Wiki-128 \mbox{(JS-div)} \textbf{perm}]{\hspace{-1em}\includegraphics[width=0.27\textwidth]{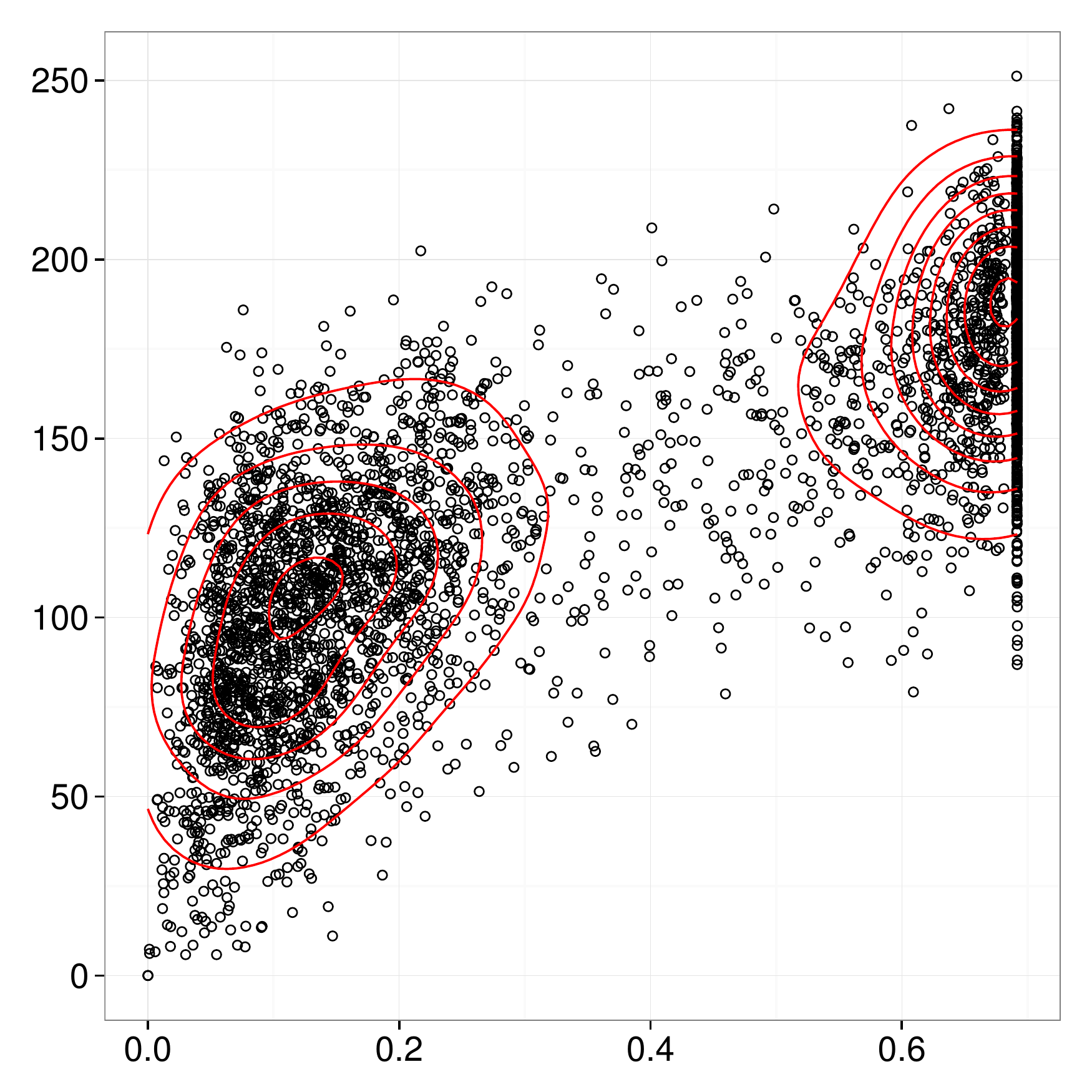}\hspace{-1em}}
\caption{\label{FigProjCorr}
Distance values in the projected space (on the y-axis) plotted against original distance values (on the x-axis). Plots \ref{CorrRandL2} and \ref{CorrRandCos} use random projections. The remaining plots rely on permutations. Dimensionality of the target space is~64.
All plots except Plot \ref{CorrRandCos} represent projections to $L_2$. In Plot \ref{CorrRandCos} the target distance function is the cosine similarity. Distances are computed for pairs of points sampled at random. Sampling is conducted from two strata: a complete subset and a set of points that are 100 nearest neighbors of randomly selected points. All data sets have one million entries.}
\end{figure*}

Images were compared using a metric function called  the
Signature Quadratic Form Distance (SQFD).
This distance is computed as a quadratic form,
where the matrix is re-computed for each pair of images using a 
heuristic similarity function applied to cluster representatives.
It is a distance metric defined over vectors from an infinite-dimensional space such that
each vector has only finite number of non-zero elements.
For further details, please, see the thesis of Beecks~\cite{Beecks:2013}. % I am not sure we need to cite Beecks:2010
SQFD was shown to be effective \cite{Beecks:2013}.
Yet, it is nearly two orders of magnitude slower compared to $L_2$.

\textbf{Wiki-sparse} is a set of four million sparse TF-IDF vectors (created via GENSIM \cite{GENSIM}).
On average, these vectors have 150 non-zero elements out of $10^5$. 
Here we use a cosine similarity,  
which is a symmetric non-metric distance: 
$$d(x,y) = 1-\left(\sum_{i=1}^n x_i y_i\right)\left(\sum_{i=1}^n x_i^2\right)^{-1/2}\left(\sum_{i=1}^n y_i^2 \right)^{-1/2}.$$
Computation of the cosine similarity between sparse vectors relies on an efficient procedure to obtain an intersection of non-zero element indices. To this end, we use an all-against-all SIMD comparison instruction as was suggested by Schlegel~et~al.~\cite{schlegel2011fast}. 
This distance function is relatively fast being only about 5x slower compared to $L_2$.
%while a straightforward implementation relying on a classic intersection algorithm (akin
%to the merge sort) is 15 times slower than $L_2$.

\textbf{Wiki-$i$}  consist of dense vectors of topic histograms created using the Latent Dirichlet Allocation (LDA)\cite{blei2003latent}.
The index $i\in\{8,128\}$ denotes the number of topics. 
To create these sets, we trained a model on one half of the Wikipedia collection and then applied it to the other half (again using GENSIM \cite{GENSIM}). Zero values were replaced by small numbers ($10^{-5}$) to avoid division by zero in the distance calculations. 
Two distance functions were used for these data sets: the  Kullback-Leibler~(KL) divergence:
$\sum_{i=1}^n   x_i \log \frac{x_i}{y_i}$
%$$
%\sum_{i=1}^n   x_i \log \frac{x_i}{y_i},
%$$
and its symmetrized version called the Jensen-Shannon (JS) divergence:
$$
d(x,y)=\frac{1}{2}\sum_{i=1}^n \left[x_i \log x_i + y_i \log y_i  - (x_i+y_i)\log \frac{x_i +y_i}{2}\right].
$$
Both the KL- and the JS-divergence are non-metric distances. Note that the KL-divergence is not even symmetric. 

Our implementation of the KL-divergence relies on the precomputation of logarithms at index time. Therefore, during retrieval it is as fast as $L_2$.
In the case of JS-divergence, it is not possible to precompute $\log (x_i+y_i)$ and, thus, it is about 10-20 times slower compared to $L_2$.

\textbf{DNA} is a collection of DNA sequences sampled from the Human Genome~\footnote{\url{http://hgdownload.cse.ucsc.edu/goldenPath/hg38/bigZips/}}.
Starting locations were selected randomly and uniformly (however, 
lines containing the symbol \texttt{N} were excluded). The length of the sequence
was sampled from $\mathcal{N}(32,4)$.
The employed distance function was the \emph{normalized Levenshtein distance}. 
This non-metric distance is equal to 
the minimum number of edit operations (insertions, deletions, substitutions),
needed to convert one sequence into another,
divided by the maximum of the sequence lengths.

\begin{figure*}
\centering
\subfloat[\label{Fig2Rand1}SIFT ($L_2$)]{\includegraphics[width=0.33\textwidth]{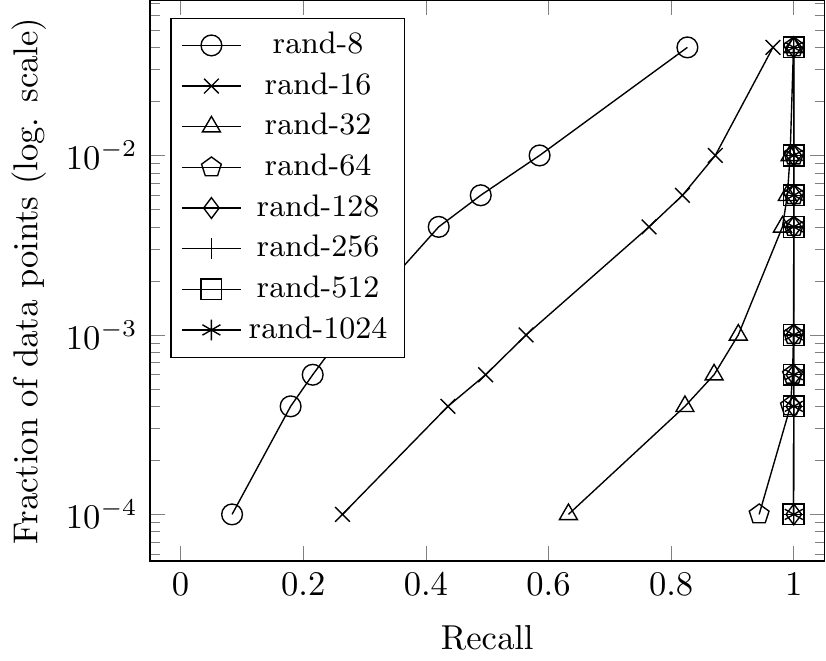}}
\subfloat[\label{Fig2Rand2}Wiki-sparse (cosine similarity)]{\includegraphics[width=0.33\textwidth]{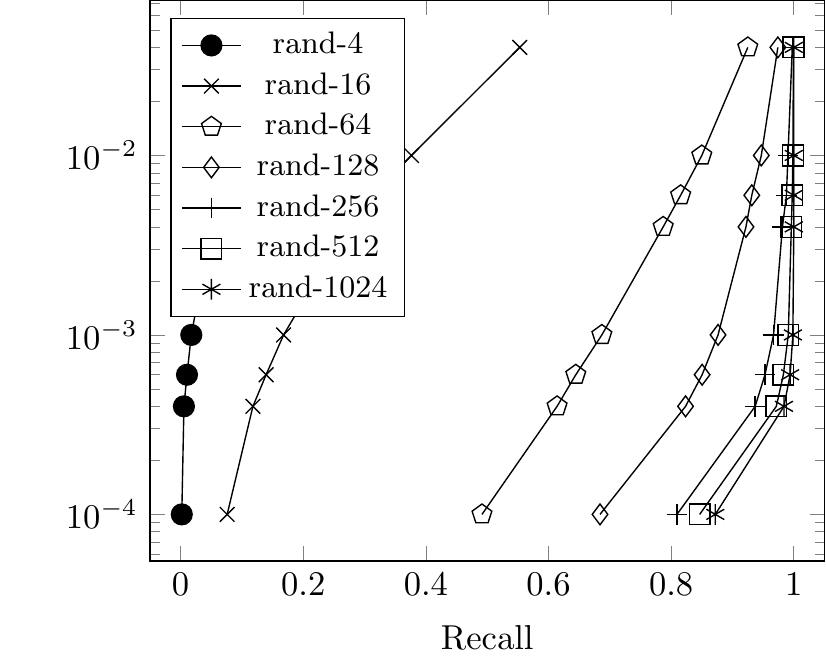}}
\subfloat[Wiki-8 (KL-divergence)]{\includegraphics[width=0.33\textwidth]{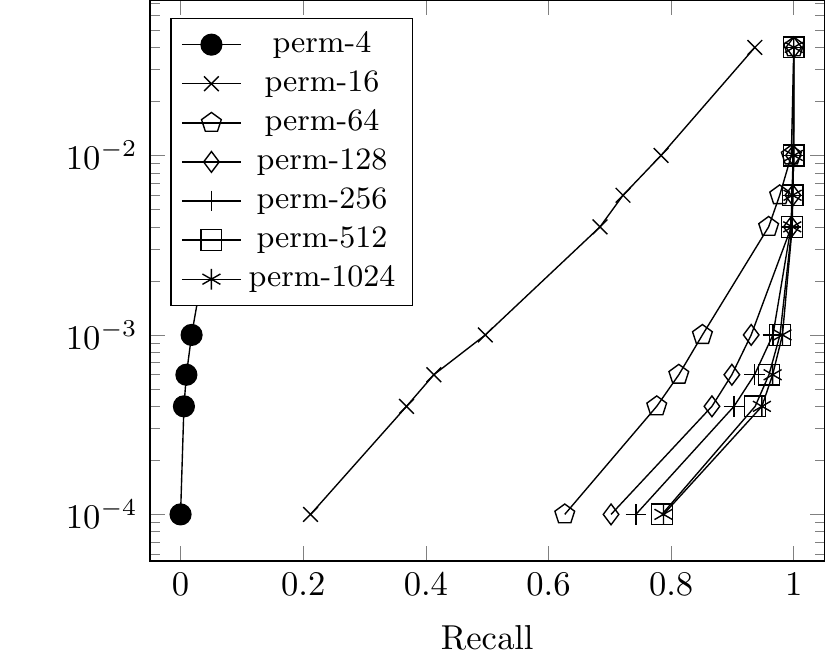}}
\\
\subfloat[SIFT ($L_2$)]{\includegraphics[width=0.33\textwidth]{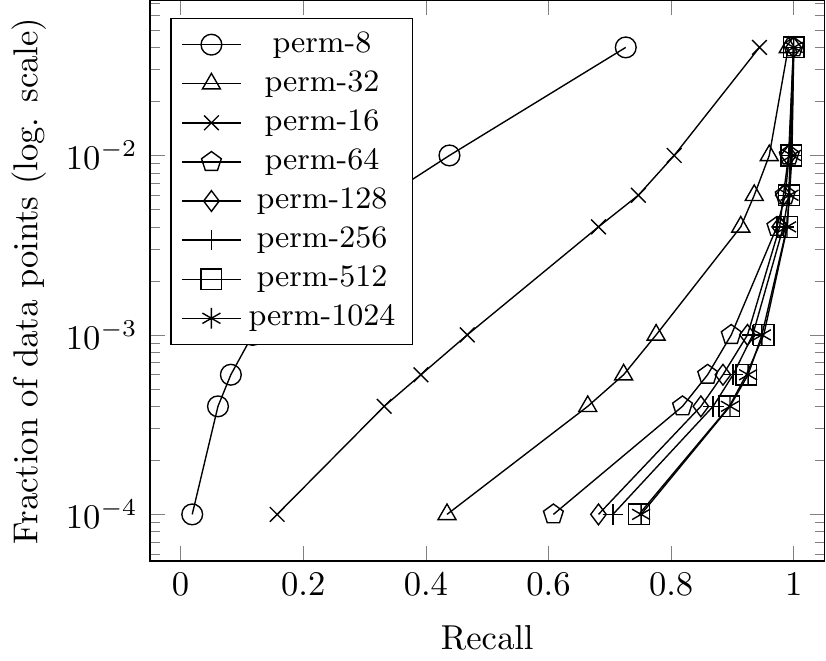}}
\subfloat[\label{Fig2BadWikiSparse}Wiki-sparse (cosine similarity)]{\includegraphics[width=0.33\textwidth]{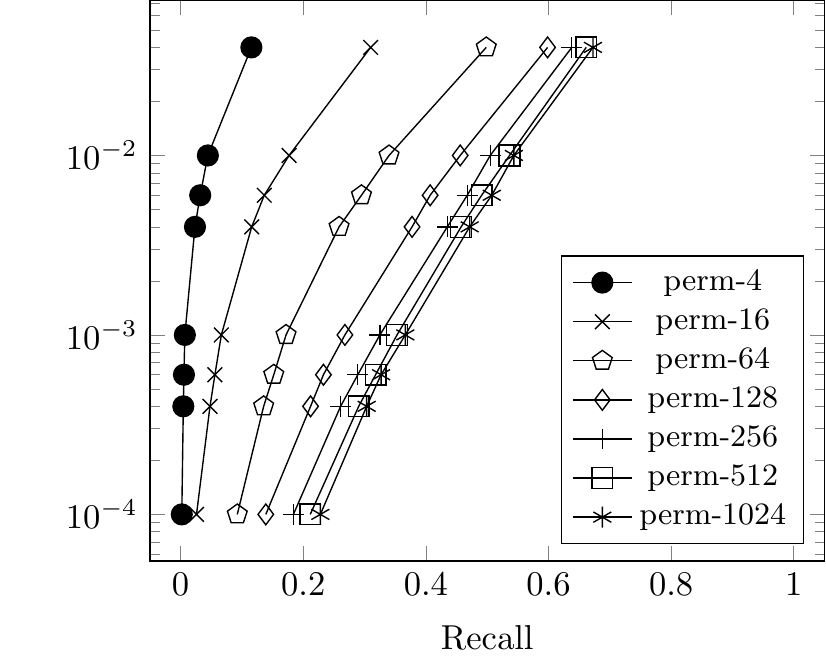}}
\subfloat[\label{Fig2BadWikiKL}Wiki-128 (KL-divergence)]{\includegraphics[width=0.33\textwidth]{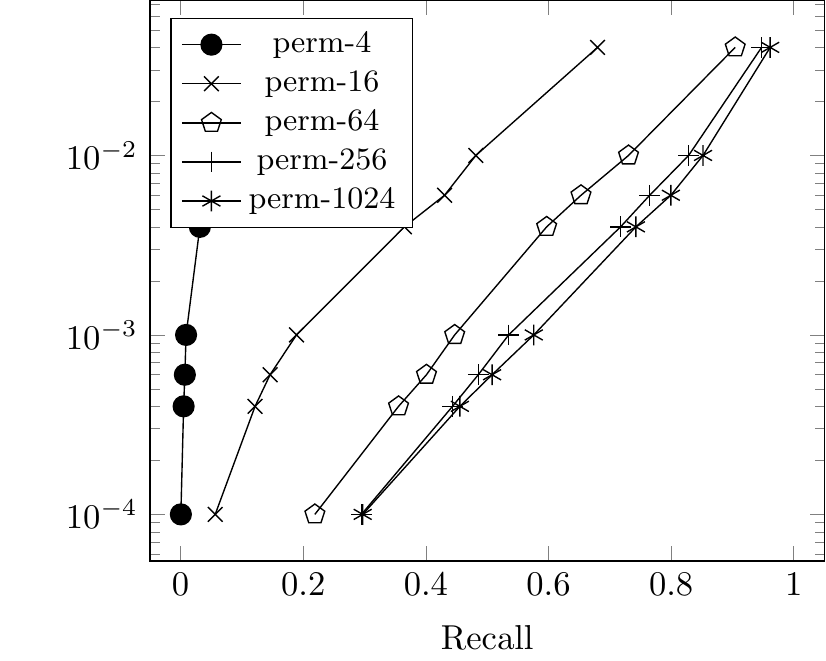}}
\\
\subfloat[DNA (Normalized Levenshtein)]{\includegraphics[width=0.33\textwidth]{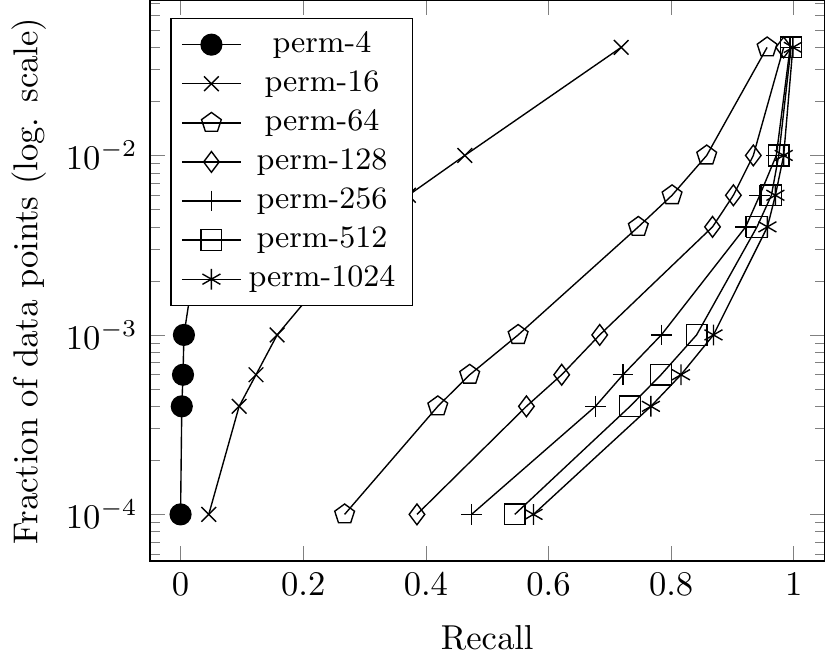}}
\subfloat[ImageNet (SQFD)]{\includegraphics[width=0.33\textwidth]{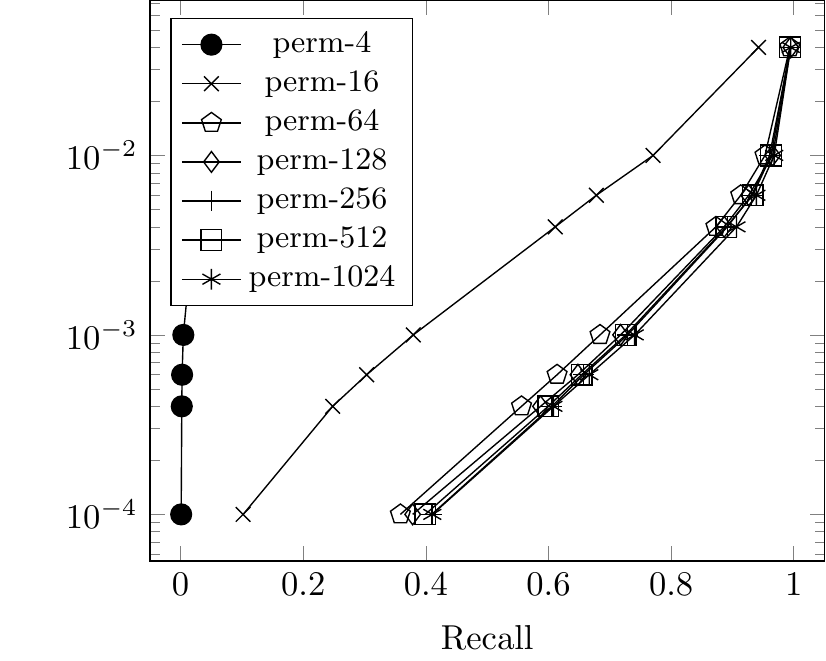}}
\subfloat[Wiki-128 (JS-divergence)]{\includegraphics[width=0.33\textwidth]{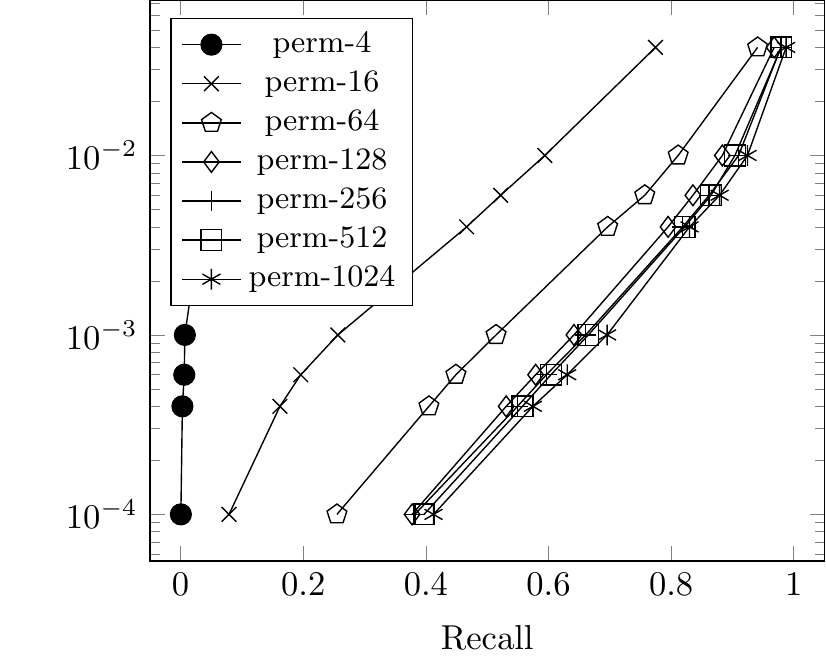}}
\caption{\label{FigProjQuality}A fraction of candidate records that are necessary to 
retrieve to ensure a desired recall level (10-NN search). The candidate entries are ranked in a projected space using either the cosine similarity (only for Wiki-sparse) or $L_2$ (for all the other data sets).
Two types of projections are used: random projections (rand) and permutations (perm).
In each plot, there are several lines that represent projections of different dimensionality.
Each data (sub)set in this experiment contains one million entries.}
\end{figure*}

\subsection{Tested Methods\label{SectionMethods}}
%We implemented and tested several permutation methods as well
%as some of the related approaches, in particular, the OMEDRANK
%heuristic due to Fagin~et~al.~\cite{Fagin2003}.
%Whenever appropriate, we explain why a method was excluded. 
Table~\ref{TableIndexSizeTime} lists all implemented methods and
provides information on index creation time and size.

\textbf{Multiprobe-LSH (MPLSH)} is implemented in the library LSHKit \footnote{Downloaded
from \url{http://lshkit.sourceforge.net/}}. It is designed to work only for $L_2$.
Some parameters are selected automatically
using the cost model proposed by Dong et al.~\cite{Dong_et_al:2008}.
However, the size of the hash table $H$, the number of hash tables $L$, and
the number of probes $T$ need to be chosen manually.
We previously found that (1) $L=50$ and $T=10$ provided a near optimal performance and
(2) performance is not affected much by small changes in $L$ and $T$ \cite{boytsov2013learning}.
This time, we re-confirmed this observation by running a small-scale grid search in the vicinity of $L=50$ and $T=50$ for $H$ equal to the number of points plus one. 
The MPLSH generates a list of candidates that are directly compared against the query.
This comparison involves the optimized SIMD implementation of $L_2$. 

\textbf{VP-tree} is a classic space decomposition tree that 
recursively divides the space with respect to
a randomly chosen pivot $\pi$\cite{yianilos1993data,uhlmann1991satisfying}.
For each partition, we compute a median value $R$ of the distance from $\pi$
to every other point in the current partition.
The pivot-centered ball with the radius $R$ is used to partition the space:
the inner points are placed into the left subtree, while the outer points
are placed into the right subtree (points that are exactly at distance $R$ from $\pi$
can be placed arbitrarily). 

Partitioning stops when the number of points
falls below the threshold $b$. The remaining points are organized in a form of a bucket.
In our implementation, all points in a bucket are stored in the same chunk of memory.
For cheap distances (e.g., $L_2$ and KL-div) this placing strategy can halve retrieval
time.

If the distance is the metric, the triangle inequality can be used to 
prune unpromising partitions as follows: imagine that $r$ is a radius
of the query and the query point is inside the pivot-centered ball (i.e.,
in the left subtree). If $R - d(\pi, q) > r$, the right partition cannot
have an answer, i.e., the right subtree can be safely pruned. 
If the query point is in the right partition, we can prune the left subtree
if $d(\pi, q) - R > r$. The nearest-neighbor search is simulated as a range
search with a decreasing radius: Each time we evaluate the distance between $q$
and a data point, we compare this distance with $r$. 
If the distance is smaller,
it becomes a new value of $r$. 
In the course of traversal,
we first visit the
closest subspace (e.g., the left subtree if the query is inside the pivot-centered ball).

For a generic, i.e., not necessarily metric, space, the pruning conditions can be modified. For example,
previously we used a liner ``stretching'' of the triangle inequality \cite{boytsov2013learning}. In this work,
we employed a simple polynomial pruner. More specifically, the right
partition can be pruned if the query is in the left partition and $(R - d(\pi,q))^\beta \alpha_{left} > r$.
The left partition can be pruned if the query is in the right partition
and $(d(\pi,q) - R)^\beta \alpha_{right} > r$.

We used $\beta=2$ for the KL-divergence and $\beta=1$ for every other distance function.
The optimal parameters $\alpha_{left}$ and $\alpha_{right}$ can be found by a trivial grid-search-like procedure with a shrinking grid step \cite{boytsov2013learning} (using a subset of data). 
%For efficiency reasons, tuning is carried out on a subset of the data set.

\textbf{\kg} (a proximity graph) is a data structure in which data points are associated
with graph nodes and $k$ edges are connected to $k$
nearest neighbors of the node.
The search algorithm relies on a concept ``the closest neighbor of my closest neighbor is my neighbor as well.'' 
This algorithm can start at an arbitrary node and recursively
transition to a neighbor point (by following the graph edge)
that is closest to the query.
This greedy algorithm stops when the current point $x$ is closer to the query
than any of the $x$'s neighbors.
However, this algorithm can be trapped in a local minima \cite{dong2011}.
Alternatively, the termination condition can be defined in
terms of an extended neighborhood \cite{sebastian2002metric,malkov2014approximate}.

Constructing an \emph{exact} \kg{} is hardly feasible for a large data set,
because, in the worst case, the number of distance computations is $O(n^2)$,
where $n$ in the number of data points. 
While there are amenable metric spaces where an exact graph can be computed 
more efficiently than in $O(n^2)$, see e.g. \cite{paredes2006practical},
the quadratic cost appear to be unavoidable in many cases, especially if the 
distance is not a metric or the intrinsic dimensionality is high.

An \emph{approximate} \kg{} can be constructed more efficiently.
In this work, we employed two different graph construction algorithms:
the NN-descent proposed by Dong et al.~\cite{dong2011efficient} and 
the search-based insertion algorithm used by Malkov et~al.~\cite{malkov2014approximate}.
The NN-descent is an iterative procedure initialized with randomly selected 
nearest neighbors. In each iteration, a random sample of queries is selected
to participate in neighborhood propagation.

Malkov et al.~\cite{malkov2014approximate} called their method a \emph{Small World} (SW) graph.
The graph-building algorithm finds an insertion point by running the same algorithm
that is used during retrieval. 
Multiple insertion attempts are carried out
starting from a random point. 

The \emph{open-source} implementation of NN-descent is publicly available online.\footnote{\url{https://code.google.com/p/nndes/}}. However, it comes without a search algorithm.
Thus, we used the algorithm due to Malkov~et~al.~\cite{malkov2014approximate},
which was available in the Non-Metric Space Library \cite{SISAP2013}.
We applied both graph construction algorithms. Somewhat surprisingly,
in all but two cases, NN-descent took (much) longer time to converge.
For each data set, we used the graph-construction algorithm that performed
better on a subset of the data.
Both graph construction algorithms are computationally expensive and are,
therefore, constructed in a multi-threaded mode (four threads).
Tuning of \kgs{} involved manual selection of two parameters $k$ and the decay coefficient (tuning was carried out on a subset of data).
The latter parameter, which is used only for NN-descent, defines the convergence speed. 

\textbf{Brute-force filtering} is a simple approach where 
we exhaustively compare the permutation of the query against permutation
of every data point.
We then use incremental sorting to select $\gamma$ permutations
closest to the query permutation. These $\gamma$ entries
represent candidate records compared directly against the query using the
original distance. 

As noted in \S~\ref{SectionLitSurvey}, 
the cost of the filtering stage is high. Thus, we use this algorithm
only for the computationally intensive distances: SQFD and the Normalized Levenshtein
distance. 
Originally, both in the case of SQFD and Normalized Levenshtein distance, 
good performance was achieved with permutations of the size 128.
However, Levenshtein distance was applied to DNA sequences,
which were strings whose average length was only 32.
Therefore, using uncompressed permutations of the size 128
was not space efficient (128 32-bit integers use 512 bytes).
Fortunately, 
we can achieve the same performance using bit-packed binary permutations 
with 256 elements, each of which requires only 32 bytes.

The optimal permutation size was found by a small-scale grid search (again using a subset of data).
Several values of $\gamma$
(understood as a fraction of the total number of points) were manually selected
to achieve recall in the range 0.85-0.9.

\textbf{NAPP} is a neighborhood approximation index described in \S~\ref{SectionLitSurvey} \cite{tellez2013succinct}.
Our implementation is different from the proposition  of 
Ch\'{a}vez et al.~\cite{Chavez_et_al:2015} and  Tellez~et~al.~\cite{tellez2009brief} 
in at least two ways: (1) we do not compress the index and (2) we use a simpler
algorithm, namely, the ScanCount, to merge posting lists \cite{li2008efficient}.
For each entry in the database, there is a counter. When we read a posting list
entry corresponding to the object $i$, we increment counter $i$. 
To improve cache utilization and overall performance, 
we split the inverted index into small chunks, 
which are processed one after another.
Before each search counters are zeroed using the function \texttt{memset} from a standard C library.

Tuning NAPP involves selection of three parameters $m$ (the total number of pivots),
$m_i$ (the number of indexed pivots), and $t$.
The latter is equal to the minimum number of 
indexed pivots that has to be shared between the query and a data point. 
By carrying out a small-scale grid search,
we found that increasing $m$ improves both recall
and decreases retrieval time, yet, improvement is small beyond $m=500$. 
At the same time, computation of one permutation entails computation of $m$ distances to pivots.
Thus, larger values of $m$ incur higher indexing cost.
Values of $m$ between 500 and 2000 provide a good trade-off.
Because the indexing algorithm is computationally expensive,
it is executed in a multi-threaded mode (four threads).

Increasing $m_i$ improves recall at the expense of retrieval efficiency:
The larger is $m_i$, the more posting lists are to be processed at query time. We found that
good results are achieved for $m_i=32$. 
Smaller values of $t$ result in high recall values. At the same time,
they also produce a larger number of candidate records, which negatively
affects performance. Thus, for cheap distances, e.g. $L_2$, we manually 
select the smallest $t$ that allows one to achieve a desired recall (using a subset of data).
For more expensive distances, 
we have an additional filtering step (as proposed by Tellez~et~al.~\cite{tellez2009brief}), 
which involves sorting by
the number of commonly indexed pivots. 

Our initial assessment showed that NAPP
was more efficient than the PP-index and at least as efficient MI-file,
which agrees with results of Ch\'{a}vez et al.~\cite{chavez2015near}.
We also compared our NAPP implementation to that of Ch\'{a}vez et al.~\cite{chavez2015near}
using the same $L_1$ data set: $10^6$ normalized CoPhIR descriptors.
At 95\% recall,  Ch\'{a}vez et al.~\cite{chavez2015near}
 achieve a 14x speed up, while we achieve a 15x speed up 
(relative to respective brute-force search implementations). 
Thus, our NAPP implementation is a competitive benchmark.
Additionally we benchmark our own implementation of Fagin~et~al.'s~OMEDRANK algorithm \cite{Fagin2003}
and found NAPP to be more efficient.
We also experimented with indexing permutations using the VP-tree, yet, this algorithm was either outperformed by the VP-tree in the original space or by NAPP. 

\subsection{Experimental Setup}
Experiments were carried out on an Linux Intel Xeon server (3.60 GHz, 32GB memory) in a single threaded mode using
the \emph{Non-Metric Space Library} \cite{SISAP2013} as an evaluation toolkit.
The code was written in C++ and compiled using GNU C++ 4.8 with the \texttt{-Ofast}
optimization flag. Additionally, we used the flag \texttt{-march=native} to enable SIMD extensions.

We evaluated performance of a 10-NN search using a procedure similar to a five-fold cross validation.
We carried out five iterations, in which a data set was randomly split into two parts.
The larger part was indexed and the smaller part comprised queries~\footnote{For cheap distances (e.g., $L_2$) the query set has the size 1000, while for more expensive ones (such as the SQFD),
we used 200 queries for each of the five splits.}.
For each split, we evaluated retrieval performance
by measuring the average retrieval time, the improvement in efficiency (compared
to a single-thread brute-force search), the recall, the index creation time, and the memory consumption.
The retrieval time, recall, and the improvement in efficiency were aggregated over five splits.
To simplify our presentation, in the case of non-symmetric KL-divergence,
we report results only the for the left queries. Results for the right
queries are similar.
%We also computed 95\% confidence intervals, but they were too narrow to be of interest.
%Thus, we do not report them.
%we ran either 200 queries (for expensive
%distances: SQFD and the normalized Levenshtein) or 1000 queries for cheap distances such as $L_2$.

Because we are interested in high-accuracy (near 0.9 recall) methods,
we tried to tune parameters of the methods (using a subset of the data)
so that their recall falls in the range 0.85-0.95. 
Method-specific tuning procedures are described in respective subsections of Section \ref{SectionMethods}.

\subsection{Quality of Permutation-Based Projections\label{SectionIndexability}}
Recall that permutation methods are filter-and-refine approaches
that map data from the original space to $L_2$ or $L_1$.
Their accuracy depends on the quality of this mapping, which we assess in this subsection. 
To this end, we explore (1) the relationship between the original
distance values and corresponding values in the projected space, (2) the
relationship between the recall and the fraction of candidate records scanned in response to a query. 

Figure \ref{FigProjCorr} shows distance values in the original space (on the x-axis)
vs. values in the projected space (on the y-axis) for eight combinations
of data sets and distance functions. Points were randomly sampled from two strata: 
 a complete subset and a set of points that are 100 nearest neighbors of randomly selected points. 
Of the presented panels, \ref{CorrRandL2} and \ref{CorrRandCos} correspond to the classic random projections. The remaining panels show permutation-based projections.

Classic random projections are known to preserve inner products and distance values \cite{bingham2001random}. 
Indeed, the relationship between the distance in the original and the projected space appears to be approximately linear in  panels \ref{CorrRandL2} and \ref{CorrRandCos}.
Therefore, it preserves the relative distance-based order of points with respect 
to a query.
For example, there is a high likelihood for the nearest neighbor in the original space 
to remain the nearest neighbor in the projected space.
In principle, any monotonic relationship---not necessarily linear--will suffice \cite{skopal2007}. 
If the monotonic relationship holds at least approximately, 
the projection typically distinguishes between points close to the query and points that are far away.

For example, the projection in panel \ref{GoodProj1} appears to be quite good, which is not
entirely surprising, because the original space is Euclidean. The projections
in panels \ref{GoodProj2} and \ref{GoodProj3} are also reasonable, but not as good as one
in panel \ref{GoodProj1}. The quality of projections in panels \ref{UnkProj1} and \ref{UnkProj2}
is somewhat uncertain. The projection in panel \ref{BadProj1}--which represents
the non-symmetric and non-metric distance--is obviously poor. Specifically, there are two
clusters: one is close to the query (in the original distance) and the other is far away.
However, in the projected space these clusters largely overlap.

Figure \ref{FigProjQuality} contains nine panels that plot 
recall (on x-axis) against a fraction of candidate records necessary to 
retrieve  to ensure this recall level (on y-axis). 
In each plot, there are several lines that represent projections of different dimensionality.
Good projections (e.g., random
projections in panels \ref{Fig2Rand1} and \ref{Fig2Rand2}) correspond to steep curves:
recall approaches one even for a small fraction of candidate records retrieved.
Steepness depends on the projection dimensionality. However, good projection curves 
are steep even in relatively low dimensions.

The worst projection according to Figure \ref{FigProjCorr} is in panel \ref{BadProj1}.
It corresponds to the Wiki-128 data set with distance measured by KL-divergence.
Panel \ref{Fig2BadWikiKL} in Figure \ref{FigProjQuality}, corresponding
to this combination of the distance and the data set, also confirms the low quality
of the projection. For example, given a permutation of dimensionality 1024,
scanning 1\% of the candidate permutations achieves roughly a 0.9 recall.
An even worse projection example is in panel \ref{Fig2BadWikiSparse}. 
In this case, regardless of the dimensionality, scanning 1\% of the 
candidate permutations achieves recall below 0.6. 

At the same time, for majority of projections in other panels, scanning 1\%
of the candidate permutations of dimensionality 1024 achieves an almost perfect recall.
In other words, for some data sets, it is, indeed, possible to obtain a small set of candidate
entries containing a true near-neighbor by searching in the permutation space.

\begin{figure*}
\centering
\includegraphics[width=1\textwidth]{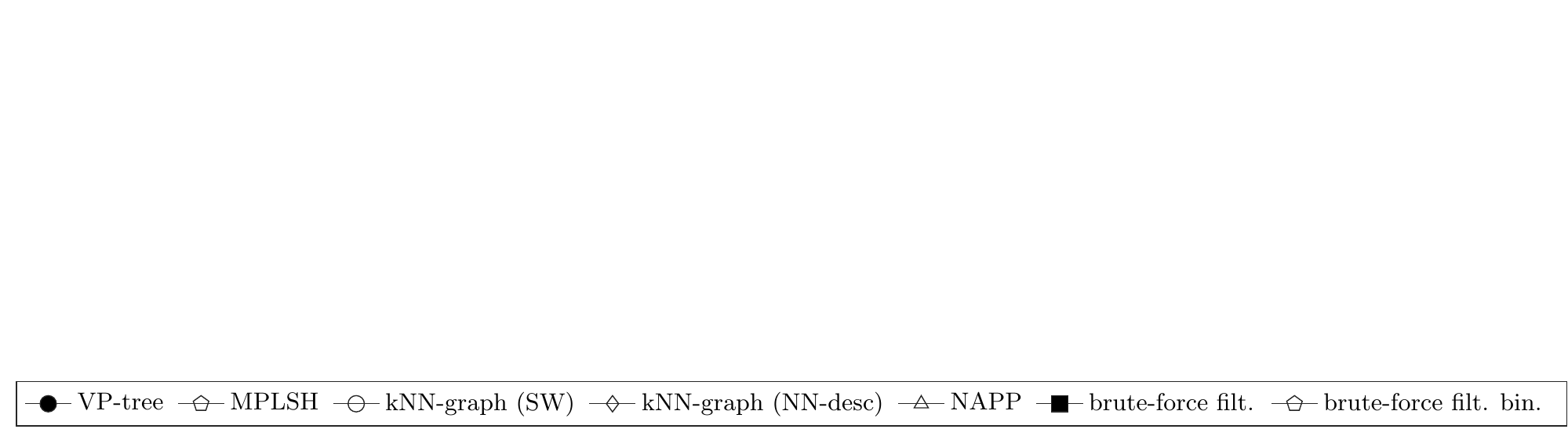}
\subfloat[\label{PanelFinalSIFT}SIFT ($L_2$)]{\includegraphics[width=0.33\textwidth]{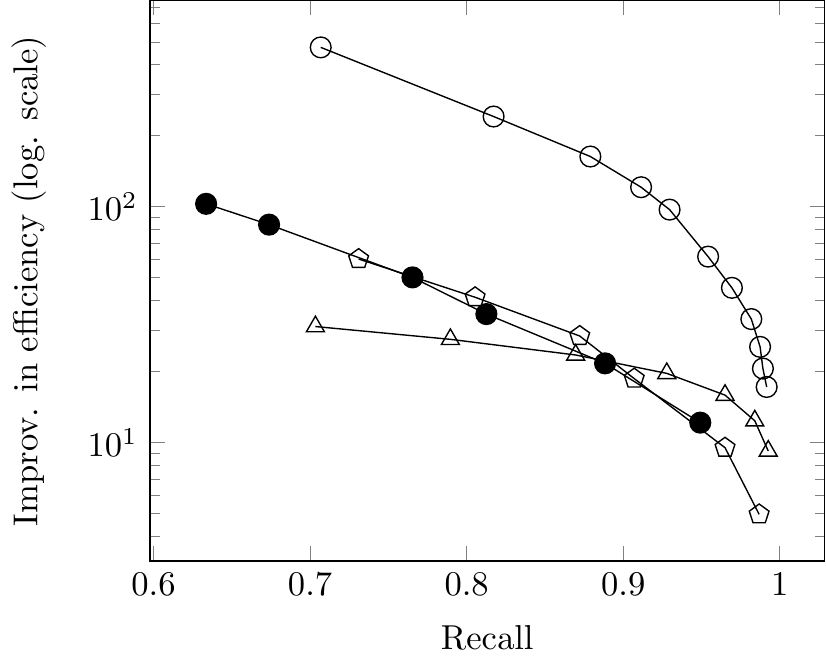}}
\subfloat[\label{PanelFinalCoPhIR}CoPhIR ($L_2$)]{\includegraphics[width=0.33\textwidth]{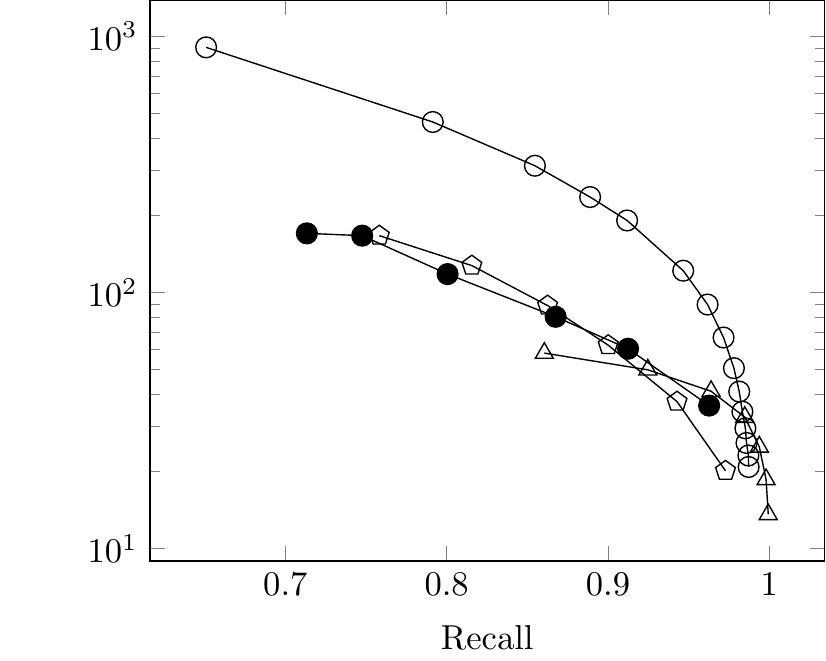}}
\subfloat[\label{FinalImageNet}ImageNet (SQFD)]{\includegraphics[width=0.33\textwidth]{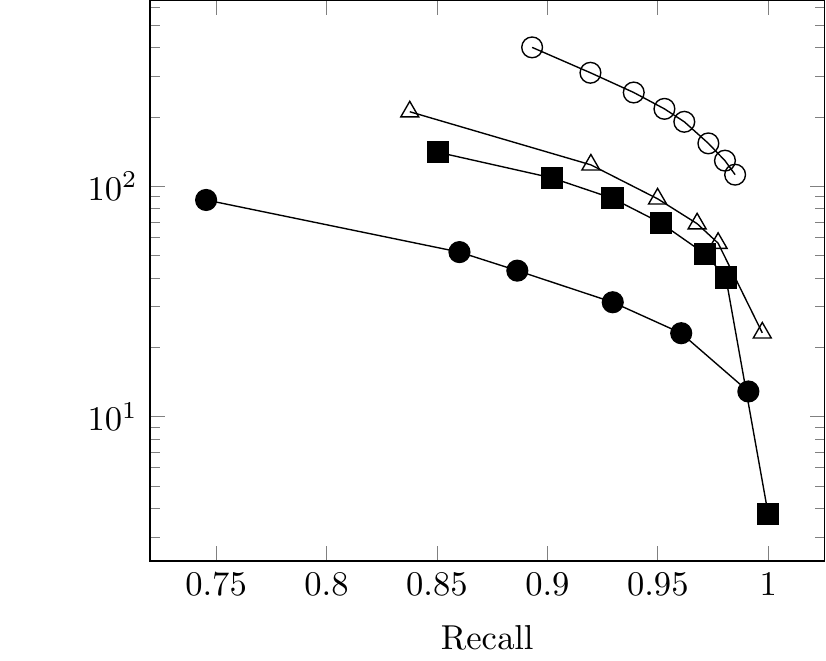}}
\\
\subfloat[\label{VPWin1}Wiki-8 (KL-div.)]{\includegraphics[width=0.33\textwidth]{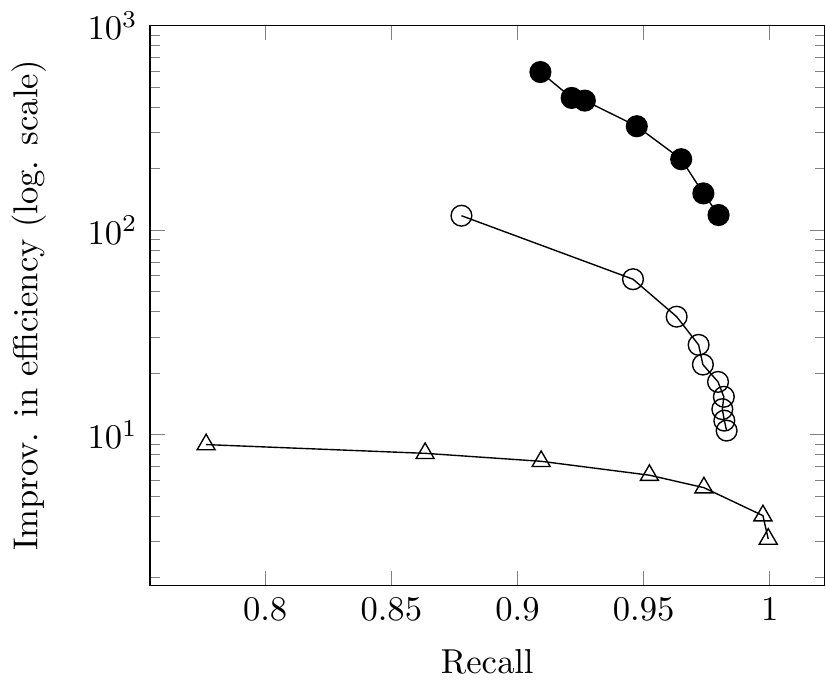}}
\subfloat[\label{VPWin2}Wiki-8 (JS-div.)]{\includegraphics[width=0.33\textwidth]{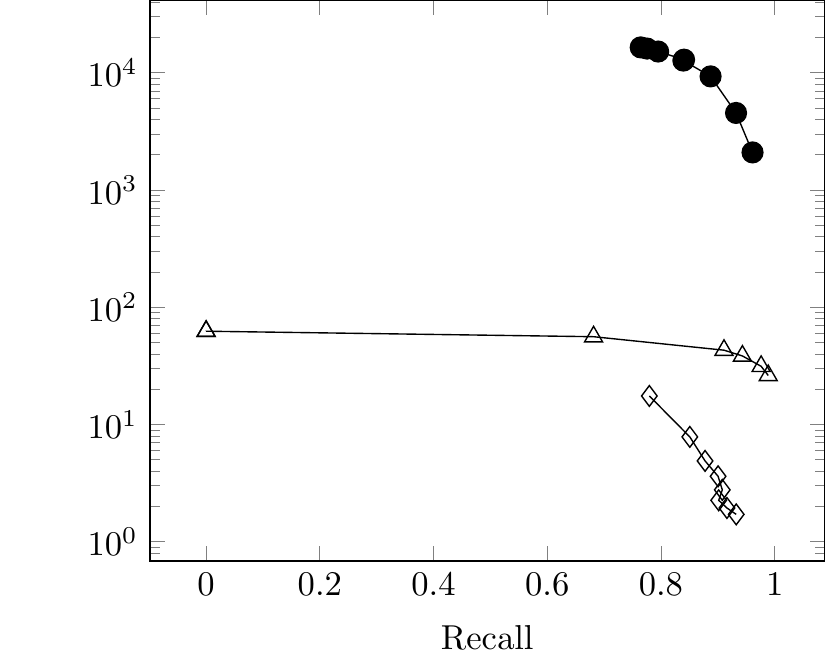}}
\subfloat[\label{BinPermWin}DNA (Normalized Levenshtein)]{\includegraphics[width=0.33\textwidth]{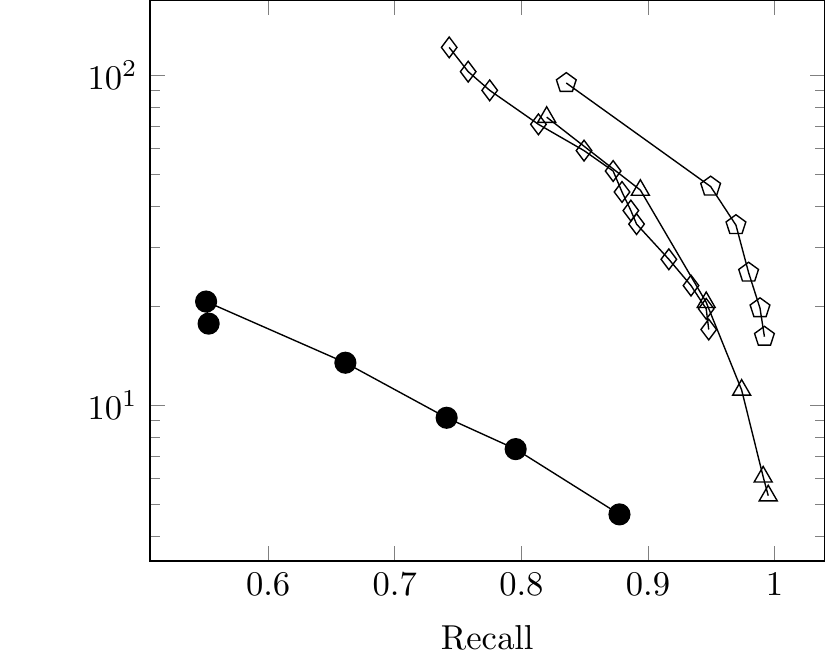}}
\\
\subfloat[Wiki-128 (KL-div.)]{\includegraphics[width=0.33\textwidth]{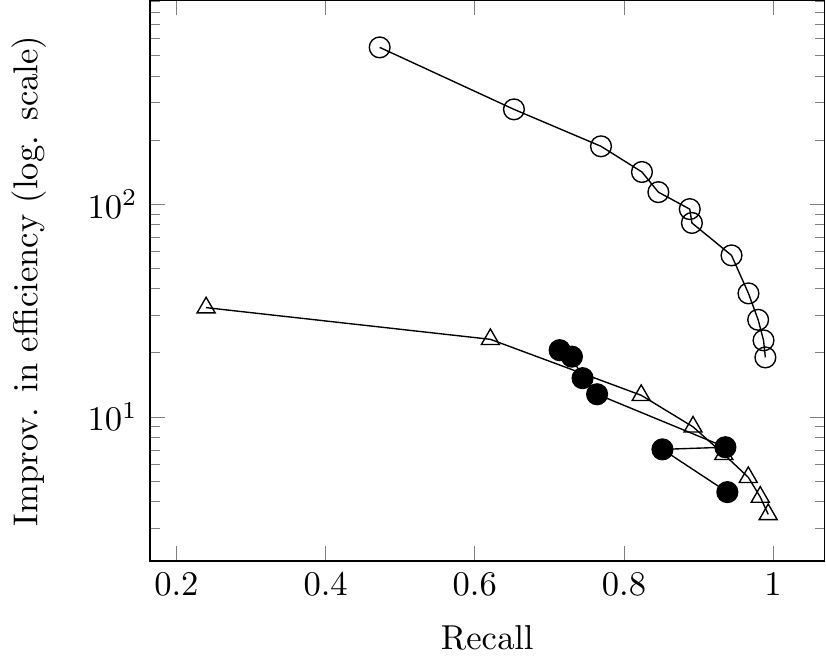}}
\subfloat[Wiki-128 (JS-div.)]{\includegraphics[width=0.33\textwidth]{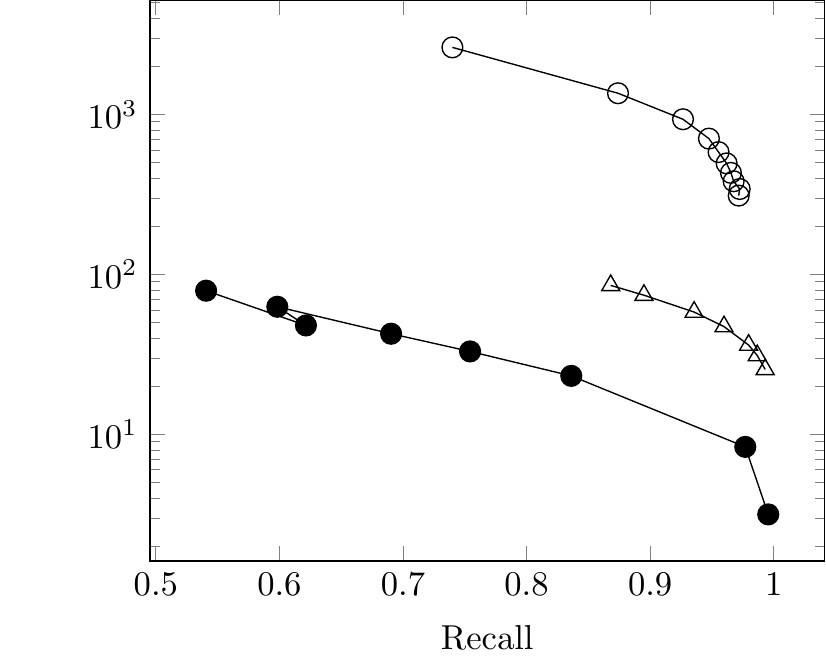}}
\subfloat[\label{PanelWikiSparse}Wiki-sparse (cosine simil.)]{\includegraphics[width=0.33\textwidth]{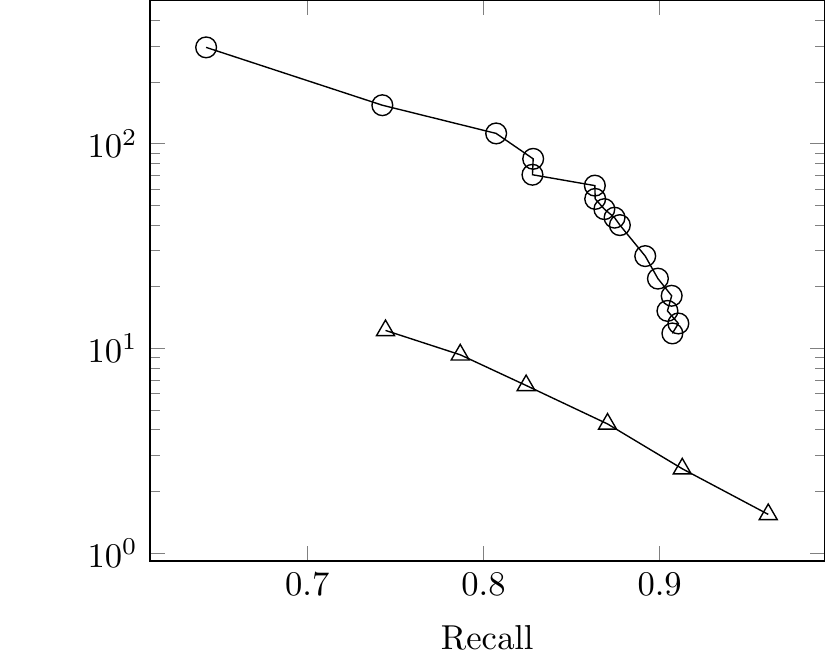}}
\caption{\label{FigEffVsRecall}Improvement in efficiency vs. recall for various data sets (10-NN search). 
Each plot includes one of the two implemented \knn graph algorithms: Small World (SW) or NN-descent (NN-desc). }
\end{figure*}

\subsection{\label{SectionMainPerf}Evaluation of Efficiency vs Recall}
In this section, we use complete data sets listed in Table \ref{TableDataSet}.
Figure \ref{FigEffVsRecall} shows nine data set specific panels 
with improvement in efficiency vs. recall. 
Each curve captures 
 method-specific results with parameter settings tuned to achieve recall in the range of 0.85-0.95.

It can be seen that in most data sets the permutation method NAPP is a competitive baseline.
In particular, panels \ref{PanelFinalSIFT} and \ref{PanelFinalCoPhIR} show NAPP outperforming
the state-of-the art implementation of the multi-probe LSH (MPLSH) for recall larger than 0.95. 
This is consistent with findings of Ch\'{a}vez et al.~\cite{chavez2015near}.

In that, in our experiments, there was no single best method. \knn graphs
substantially outperform other methods in 6 out of 9 data sets.
However, in low-dimensional data sets shown in panels \ref{VPWin1} and \ref{VPWin2},
the VP-tree outperforms the other methods by a wide margin.
The Wiki-sparse data set (see panel \ref{PanelWikiSparse}), which has high representational dimensionality, is quite challenging.
Among implemented methods, only \knn graphs are efficient for this set.

Interestingly, the winner in panel \ref{BinPermWin} is a brute-force filtering using
binarized permutations. Furthermore, the brute-force filtering 
is also quite competitive in panel \ref{FinalImageNet},
where it is nearly as efficient as NAPP. 
In both cases, the distance function is computationally intensive
and a more sophisticated permutation index does not offer 
a substantial advantage over a simple brute-force search in the permutation space.

Good performance of \knn graphs comes at the expense of long indexing time.
For example, it takes almost four hours to built the index for the Wiki-sparse data set
using as many as four threads (see Table~\ref{TableIndexSizeTime}). 
In contrast, it takes only 8 minutes in the case of NAPP (also using four threads).
In general, the indexing algorithm of \knn graphs is substantially slower than
the indexing algorithm of NAPP: it takes up to an order of magnitude longer to build 
a \knn graph. One exception is the case of Wiki-128 where the distance is the JS-divergence.
For both NAPP and \knn graph, the indexing time is nearly 40 minutes. 
However, the \knn graph retrieval is an order of magnitude more efficient.

Both NAPP and the brute-force searching of permutations have high indexing costs compared to the VP-tree.
This cost is apparently dominated by time necessary to compute permutations.
Recall that obtaining a permutation entails $m$ distance computations. 
Thus, building an index entails $N \cdot m$ distance computations, where $N$ is the number of data points. 
In contrast, building the VP-tree requires roughly $N \cdot \log_2 N/b$ distance computations, where $b$ is the size of the bucket.
In our setup, $m > 100$ while $\log_2 N/b < 20$. Therefore, the indexing step of permutation methods is typically much longer
than that of the VP-tree.

Even though permutation methods may not be the best solutions when both data and the index are kept in main memory,
they can be appealing in the case of disk-resident data \cite{amato2014mi}
or data kept in a relational database. 
Indeed, as noted by Fagin~et~al.~\cite{Fagin2003}, indexes based on the inverted files are \emph{database friendly},
because they require neither complex data structures nor many random accesses. 
\footnote{The brute-force filtering of permutations is a simpler approach, which is also database friendly.}
Furthermore, deletion and addition of records can be easily implemented.
In that, it is rather challenging to implement a dynamic version of the VP-tree on top of a relational database.

We also found that all evaluated methods perform reasonably well in the surveyed non-metric spaces.
This might indicate that there is some truth to the two folklore wisdoms:
(1) ``the closest neighbor of my closest neighbor is my neighbor as well'',
(2) ``if one point is close to a pivot, but another is far away, 
such points cannot be close neighbors''. Yet, these wisdoms are not universal.
For example, they are violated in one dimensional space with the ``distance''
$e^{-|x-y|}|x-y|$. In this space, points 0 and 1 are distant. However, we can 
select a large positive number that can be arbitrarily close to both of them,
which results in violation of properties (1) and (2).

It seems that such a paradox does not manifest  in the surveyed non-metric spaces.
In the case of continuous functions, there is non-negative
strictly monotonic transformation \mbox{$f(x)\ge 0$}, $f(0)=0$ such that $f(d(x,y))$ is 
a $\mu$-defective distance function. Thus, the distance satisfies the following inequality:
\begin{equation}\label{Eq1}
|f(d(q, a)) - f(d(q, b))| \le \mu f(d(a, b)), \mu > 0
\end{equation}
Indeed, a monotonic transformation of the cosine similarity is 
the metric function (i.e, the angular distance) \cite{van2012metric}.
The square root of the JS-divergence is metric function called
Jensen-Shannon distance \cite{endres2003new}.
The square root of all Bregman divergences (which
include the KL-divergence) is $\mu$-defective as well \cite{abdullah2012approximate}.
The normalized Levenshtein distance is a non-metric distance.
However, for many realistic data sets, the triangle inequality is rarely violated.
In particular, we verified that this is the case of our data set.
The normalized Levenshtein distance is approximately metric and, thus, it is approximately
$\mu$-defective (with $\mu=1$).

If Inequality (\ref{Eq1}) holds,
due to properties of $f(x)$, \mbox{$d(a,b)= 0$} and \mbox{$d(q,a)=0$}
implies $d(q,b)=0$. Similarly if $d(q,b)=0$, but 
$d(q,a) \ne 0$, $d(a,b)$ cannot be zero either.
Moreover, for a sufficiently large $d(q,a)$ and small $d(q,b)$, 
$d(a,b)$ cannot be small.
Thus, the two folklore wisdoms are true if the strictly monotonic distance transformation is  $\mu$-defective.

\section{Conclusions}
\label{SectionConclusion}
We benchmarked permutation methods for approximate $k$-nearest neighbor search
for generic spaces where both data and indices are stored in main memory (aiming for high-accuracy retrieval).
We found these filter-and-refine methods to be reasonably efficient.
The best performance is achieved either by NAPP or by brute-force filtering of permutations.
For example,
NAPP can outperform the multi-probe LSH in $L_2$. However, permutation methods can be
outstripped by either VP-trees or \knn graphs, partly because the filtering
stage can be costly. 

We believe that permutation methods are most useful in non-metric spaces of moderate dimensionality when:
(1) The distance function is expensive (or the data resides on disk);
(2) The indexing costs of \knn graphs are unacceptably high;
%and permutation methods provide a better alternative; 
(3) There is a need
for a simple, but reasonably efficient, implementation that operates on top of a relational database.

%ACKNOWLEDGMENTS are optional
\section{Acknowledgments}
This work was partially supported  by the iAd Center~\footnote{\url{http://www.iad-center.com/}}
and the Open Advancement of Question Answering Systems (OAQA) group~\footnote{\url{http://oaqa.github.io/}}.

We also gratefully acknowledge help of several people. 
In particular, we are thankful to Anna Belova for helping edit the experimental and concluding sections.
We thank Christian Beecks for answering questions regarding the Signature Quadratic Form Distance (SQFD) \cite{Beecks:2013};
Daniel Lemire for providing the implementation of the SIMD intersection algorithm;
Giuseppe Amato and Eric S. Tellez for help with data sets;
Lu Jiang\footnote{\url{http://www.cs.cmu.edu/~lujiang/}} for the helpful discussion of image retrieval algorithms
and for providing useful references.

We thank Nikita Avrelin and Alexander Ponomarenko for porting their proximity-graph based retrieval algorithm
to the Non-Metric Space Library\footnote{\url{github.com/searchivarius/nmslib}}. 
The results of the preliminary evaluation were published elsewhere \cite{ponomarenko2014comparative}. 
In the current publication, 
we use improved versions of the NAPP and baseline methods.
In particular,
we improved the tunning algorithm of the VP-tree
and we added another implementation of the proximity-graph based retrieval \cite{dong2011efficient}.
Furthermore, we experimented with a more diverse collection of (mostly larger) data sets. 
In particular, because of this, we found that proximity-based retrieval may not be an optimal solution in all cases, e.g., when the distance function is expensive to compute. 

%\bibliographystyle{abbrv}
%\small
%\bibliography{perm2015} 

%APPENDIX is optional.
% ****************** APPENDIX **************************************
% Example of an appendix; typically would start on a new page
%pagebreak

\normalsize

\end{document}